\documentclass{article}

\usepackage[preprint,nonatbib]{neurips_2022}

\usepackage[utf8]{inputenc} % allow utf-8 input
\usepackage[T1]{fontenc}    % use 8-bit T1 fonts
\usepackage{hyperref}       % hyperlinks
\hypersetup{colorlinks=true,citecolor=blue,linkcolor=blue, urlcolor=blue}
\usepackage{url}            % simple URL typesetting
\usepackage{booktabs}       % professional-quality tables
\usepackage{amsfonts}       % blackboard math symbols
\usepackage{nicefrac}       % compact symbols for 1/2, etc.
\usepackage{microtype}      % microtypography
\usepackage{algorithmic}
\usepackage{algorithm}
\usepackage{graphicx}
\usepackage{amsmath,bm}
\usepackage{amsthm}

\DeclareMathOperator*{\argmin}{arg\,min}
\newcommand{\abs}[1]{\left\lvert#1\right\rvert}

\usepackage{subfigure}
\usepackage{color}
\usepackage{multirow}
\usepackage{balance}
\usepackage{enumitem}

\usepackage{wrapfig}
\usepackage[english]{babel}
\usepackage{graphicx}

\usepackage{blindtext}

\usepackage{xcolor}         % colors

\def\y{{\boldsymbol y}}

\def\LL{{\mathcal L}}

\title{Bayesian Robust Graph Contrastive Learning}

\author{%
  Yancheng Wang, Yingzhen Yang\thanks{yingzhenyang.com} \\
  School of Computing and Augmented Intelligence\\
  Arizona State University\\
  Tempe, AZ 85281 \\
  \texttt{\{ywan1053,yingzhen.yang\}@asu.edu}
}

\begin{document}

\maketitle

\begin{abstract}
Graph Neural Networks (GNNs) have been widely used to learn node representations and with outstanding performance on various tasks such as node classification. However, noise, which inevitably exists in real-world graph data, would considerably degrade the performance of GNNs revealed by recent studies. In this work, we propose a novel and robust method, Bayesian Robust Graph Contrastive Learning (BRGCL), which trains a GNN encoder to learn robust node representations. The BRGCL encoder is a completely unsupervised encoder. Two steps are iteratively executed at each epoch of training the BRGCL encoder: (1) estimating confident nodes and computing robust cluster prototypes of node representations through a novel Bayesian nonparametric method; (2) prototypical contrastive learning between the node representations and the robust cluster prototypes. Experiments on public and large-scale benchmarks demonstrate the superior performance of BRGCL and the robustness of the learned node representations. The code of BRGCL is available at \url{https://github.com/BRGCL-code/BRGCL-code}.
\end{abstract}

\section{Introduction}
Graph Neural Networks (GNNs) have become popular tools for node representation learning in recent years~\cite{kipf2017semi,bruna2013spectral,hamilton2017inductive,xu2018powerful}. Most prevailing GNNs ~\cite{kipf2017semi, zhu2020simple} leverage the graph structure and obtain the representation of nodes in a graph by utilizing the features of their connected nodes.
Benefiting from such propagation mechanism, node representations obtained by GNN encoders have demonstrated superior performance on various downstream tasks such as semi-supervised node classification and node clustering.

%As a generalization of neural networks for graphs, GNNs are also likely to have poor performance trained on noisy data.
Although GNNs have achieved great success in node representation learning, current GNN approaches do not consider the noise in the input graph.
However, noise inherently exists in the graph data for many real-world applications. Such noise may be present in node attributes or node labels, which forms two types of noise, the attribute noise and the label noise.
The graph with noisy labels and attributes can significantly degrade the performance of the GNN encoders. Recent works, such as \cite{patrini2017making}, have evidenced that noisy inputs hurts the generalization capability of neural networks. Moreover, noise in a subset of the graph data can easily propagate through the graph topology to corrupt the remaining nodes in the graph data. Nodes that are corrupted by noise or falsely labeled would adversely affect the representation learning of themselves and their neighbors.

While manual data cleaning and labeling could be remedies to the consequence of noise,  they are expensive processes and difficult to scale, thus not able to handle almost infinite amount of noisy data online. Therefore, it is crucial to design a robust GNN encoder which could make use of noisy training data while circumventing the adverse effect of noise. In this paper, we propose a novel and robust method termed Bayesian Robust Graph Contrastive Learning (BRGCL) to improve the robustness of node representations for GNNs. Our key observation is that there exist a subset of nodes which are confident in their class/cluster labels. Usually, such confident nodes are far away from the class/cluster boundaries, so these confident nodes are trustworthy and noise in these nodes would not degrade the value of these nodes in training a GNN encoder. To infer such confident nodes, we propose a novel algorithm named Bayesian nonparametric Estimation of Confidence (BEC). Since the BRGCL encoder is completely unsupervised, it first infers pseudo labels of all the nodes with a Bayesian nonparametric method only based on the input node attributes, without knowing the ground truth labels or the ground truth class number in the training data. Then, BEC is used to estimate the confident nodes based on the pseudo labels and the graph structure. The robust prototype representations, as the cluster centers of the confident nodes, are computed and used to train the BRGCL encoder with a loss function for prototypical constrastive learning. The confident nodes are updated during each epoch of the training of the BRGCL encoder, so the robust prototype representations are also updated accordingly.

\subsection{Contributions}
Our contributions are as follows.

First, we propose Bayesian Robust Graph Contrastive Learning (BRGCL) where a \textit{fully unsupervised} encoder is trained on noisy graph data. The fully unsupervised BRGCL encoder is trained only on the input node attributes without ground truth labels or even the ground truth class number in the training data.  GRGCL leverages confident nodes, which are estimated by a novel algorithm termed Bayesian nonparametric Estimation of Confidence (BEC), to harvest noisy graph data without being compromised by the noise. Experimental results on popular and large-scale graph datasets evidence the advantage of BRGCL over competing GNN methods for node classification and node clustering on noisy graph data.

Second, our study reveals the importance of confident nodes in training GNN encoders on noisy graph data, which opens the door for future research in this direction. The visualization results show that the confident nodes estimated by BEC are usually far away from the class/cluster boundaries, so are the robust prototype representations. As a result, the BRGCL encoder trained with such robust prototypes is not vulnerable to noise, and it even outperforms GNNs trained with ground truth labels.

\section{Related Works}
\textbf{Graph Neural Networks.}
Graph neural networks (GNNs) have become popular tools for node representation learning. They have shown superior performance in various graph learning tasks, such as node classification, node clustering and graph classification. Given the difference in the convolution domain, current GNNs fall into two classes. The first class features spectral convolution \cite{bruna2013spectral,kipf2017semi},  and the second class \cite{hamilton2017inductive,velivckovic2017graph,xu2018powerful} generates node representations by sampling and propagating features from neighborhood. GNNs such as ChebNet~\cite{bruna2013spectral} perform convolution on the graph Fourier transforms termed spectral convolution. Graph Convolutional Network (GCN)~\cite{kipf2017semi} further simplifies the spectral convolution~\cite{bruna2013spectral} by its first-order approximation. GNNs such as GraphSAGE~\cite{hamilton2017inductive} propose to learn a function that generates node representations by sampling and propagating features from a node’s connected neighborhood to itself. Various designs of the propagation function have been proposed. For instance, Graph Attention Network (GAT)~\cite{velivckovic2017graph} proposes to learn masked self-attention layers that enable nodes to attend over their neighborhoods' features. Different from GNNs based on spectral convolution, such methods can be trained on mini-batches~\cite{hamilton2017inductive, xu2018powerful}, so they are more scalable to large graphs.

However, as pointed out by \cite{dai2021nrgnn}, the performance of GNNs can be easily degraded by noisy training data~\cite{nt2019learning}. Moreover, the adverse effects of noise in a subset of nodes can be exaggerated by being propagated to the remaining nodes through the network structure, exacerbating the negative impact of noise.

% \textbf{Node Representation Leanring}
\textbf{Existing Methods Handing Noisy Data.}
Previous works~\cite{zhang2021understanding} have shown that deep neural networks usually generalize badly when trained on input with noise. Existing literature of robust learning with noisy inputs mostly focuses on image or text domain. Such robust learning methods fall into two categories. The first category \cite{patrini2017making,goldberger2016training} mitigates the effects of noisy inputs by correcting the computation of loss function, known as loss corruption. The second category aims to select clean samples from noisy inputs for the training~\cite{malach2017decoupling,jiang2018mentornet,yu2019does,li2020dividemix, Han2018NIPS}, known as sample selection.
Recent loss correction methods usually directly modify class probabilities obtained by the models to compensate for the misguidance by the noisy inputs.  For example, \cite{goldberger2016training} corrects the predicted probabilities with a corruption matrix computed on a clean set of inputs.
On the other hand, recent sample selection methods usually select a subset of training data to perform robust learning.
Among the existing loss correction and sample selection methods, Co-teaching~\cite{Han2018NIPS} is promising which trains two deep neural networks and performs sample selection in training batch by comparing predictions from the two networks. However, such sample selection strategy does not generalize well in graph domain \cite{dai2021nrgnn} due to extraordinarily small size of labeled nodes. More details are to be introduced in Section~\ref{section:BEC}. Self-Training~\cite{li2018deeper} finds nodes with the most confident pseudo labels, and it augmented the labeled training data by incorporating confident nodes with their pseudo labels into the existing training data. However, the confident pseudo labels are not reliable because the pseudo labels are produced by a model trained on noisy data.

In addition to the above two categories of robust learning methods, recent studies \cite{kang2019decoupling, zhong2021improving, wang2021decoupling} show that decoupling the feature representation learning and the training of the classifier can also improve the robustness of the learned feature representation.

\begin{figure*}
  \centering
    \includegraphics[width=0.9\textwidth]{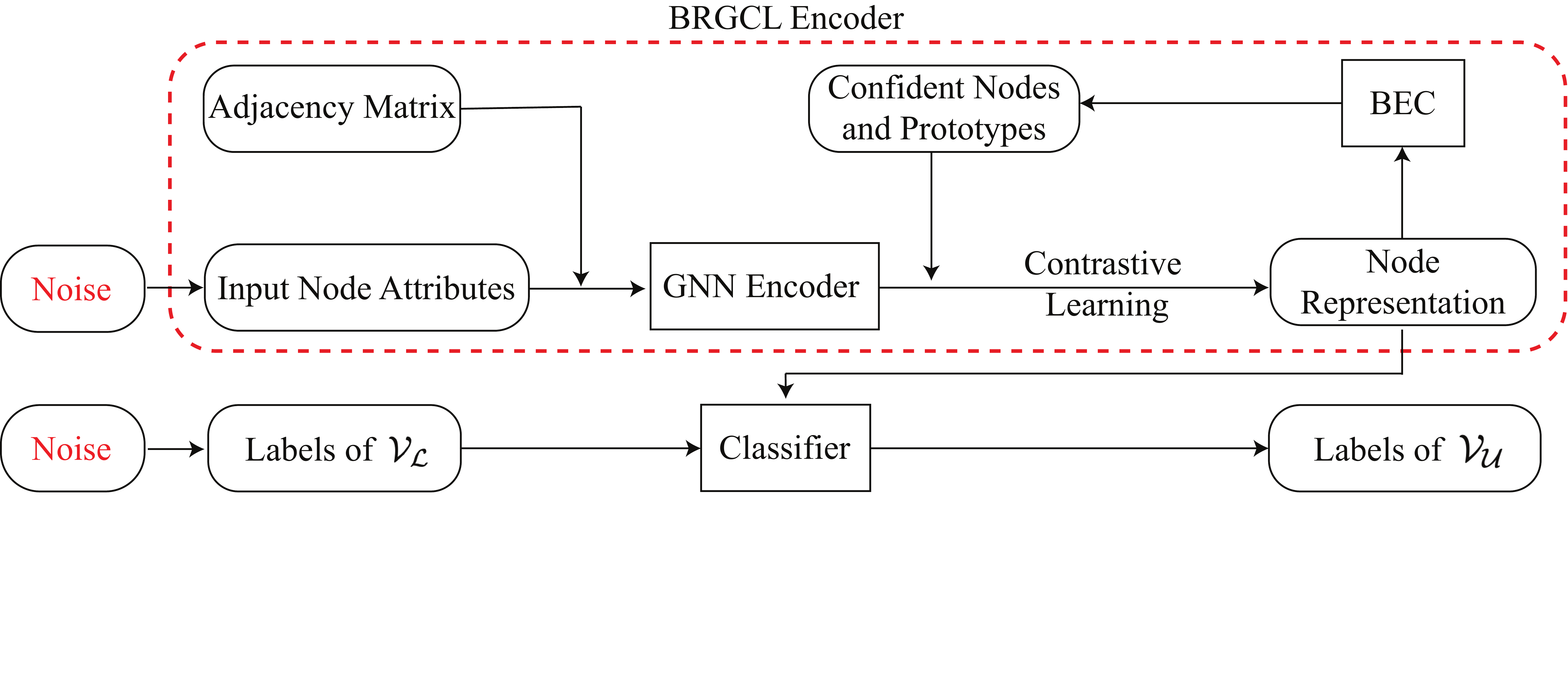}
    \vspace{-0.1cm}
  \caption{Illustration of the BRGCL encoder. BEC stands for the Bayesian nonparametric Estimation of Confidence algorithm to be introduced in Section~\ref{section:BEC}, a Bayesian nonparametric algorithm to estimate the confident nodes and their prototypes. }
  \label{fig:overall_framework}
\end{figure*}
\section{Problem Setup}
\label{sec:setup}
An attributed graph consisting of $N$ nodes is formally represented by $\mathcal{G} = (\mathcal{V}, \mathcal{E}, \mathbf{X})$, where $\mathcal V = \{v_1, v_2, \dots, v_N\}$ and $\mathcal{E}\subseteq\mathcal{V}\times\mathcal{V}$ denote the set of nodes and edges respectively. $\mathbf{X} \in \mathbb R^{N \times D}$ are the node attributes, and the attributes of each node is in $\mathbb R^d$. Let $\mathbf{A} \in \{0, 1\}^{N \times N}$ be the adjacency matrix of graph $\mathcal{G}$, with $\mathbf{A}_{ij} = 1$ if and only if $(v_i, v_j)\in \mathcal{E}$.  Let $\mathcal{V_L}$ and $\mathcal{V_U}$ denote the set of labeled nodes and unlabeled nodes respectively.

Noise usually exists in the input node attributes or labels of real-world graphs, which degrades the quality of the node representation obtained by common GCL encoders and affects the performance of the classifier trained on such representations. We aim to obtain node representations robust to noise in two cases, where noise is present in either the labels of $\mathcal{V_L}$ or in the input node attributes $\mathbf{X}$. That is, we consider either noisy label or noisy input node attributes.

The goal of BRGCL is to learn a node representation $\mathbf{H} = g(\mathbf{X},\mathbf{A})$, such that the node representations $\left\{\mathbf h_i\right\}_{i=1}^N$ are robust to noise in the above two cases, where $g(\cdot)$ is the BRGCL encoder and $\mathbf h_i$ is the $i$-th row of $\mathbf{H}$. To evaluate the performance of the robust node representations by BRGCL, the node representations $\left\{\mathbf h_i\right\}_{i=1}^N$ are used for the following two tasks.
\begin{itemize}[leftmargin=.1in]
\setlength{\itemindent}{.13in}
\item[(1)] Semi-supervised node classification, where a classifier on $\mathcal{V_L}$, and then the classifier predicts the labels of the remaining unlabeled nodes, $\mathcal{V_U}$.
% and $\mathbf{y}_i=[y_{i1},...,y_{iK} ]^{T}$ denote the one hot label of node $v_i$, where $K$ is the number of classes.

\item[(2)] Node clustering, where K-means clustering is performed on the node representations $\left\{\mathbf h_i\right\}_{i=1}^N$ to obtain node clusters.
\end{itemize}

\textbf{Notations.} Throughout this paper we use $\|\cdot\|_2$ to denote the Euclidean norm of a vector, and $[n]$ to denote all the natural numbers beween $1$ and $n$ inclusively.

%In this paper, we study the problem of semi-supervised node classification on graphs with label noise and feature noise. We aim to learn a node representation as well as a classifier that are robust to noise in labels and features of the graph data.

% However, when class labels in $\mathcal{V_L}$ are corrupted with label noise, the classification would cause the GNN training to overfit incorrect labels, and in turn lead to degraded classification performance. In addition, noise in input features will also be broadcast in the graph during the training of GNN encoders. Therefore, in our work, we aim to train a robust GNN encoder to generate node representation that is less sensitive to corruption on input feature and labels.

% \subsection{Decoupling Node Representation and Classification}
% In previous node classification models, the classifier weights are usually trained jointly with the model parameters for extracting the node representation by minimizing the cross-entropy loss between the ground truth and prediction of labeled nodes. This is also a typical pipeline for node classification with label noise. However, in this procedure, especially with a high noise ratio, the training for the classifier would greatly degrade the node representation learning. In this section, we consider decoupling the representation learning from the classification to mitigate the effects of label noise for semi-supervised node classification.
\section{Bayesian Robust Graph Contrastive Learning}

We propose Bayesian Robust Graph Contrastive Learning (BRGCL) in this section to improve the robustness of node representations. First, we review preliminaries of graph contrastive learning. Next, we propose Bayesian nonparametric Estimation of Confidence (BEC) algorithm to estimate robust nodes and prototypes. Then, we show details of node classification and node clustering. At last, we propose a decoupled training pipeline of BRGCL for semi-supervised node classification. Figure~\ref{fig:overall_framework} illustrates the overall framework of our proposed BRGCL.
\subsection{Preliminary of Graph Contrastive Learning}

The general node representation learning aims to train an encoder $g(\cdot)$, which is a two-layer Graph Convolution Neural Network (GCN) \cite{kipf2017semi}, to generate discriminative node representations. In our work, we adopt contrastive learning to train the BRGCL encoder $g(\cdot)$. To perform contrastive learning, two different views, denoted as $G^{1} = (\mathbf{X}^{1}, \mathbf{A}^{1})$ and $G^{2} = (\mathbf{X}^{2}, \mathbf{A}^{2})$ are generated through a variety of data augmentations. The representation of two generated views are denoted as $\mathbf{H}^{1}=g(\mathbf{X}^{1}, \mathbf{A}^{1})$ and $\mathbf{H}^{2}=g(\mathbf{X}^{2}, \mathbf{A}^{2})$, with $\mathbf{h}^1_i$ and $\mathbf{h}^2_i$ being the $i$-th row of $\mathbf{H}^{1}$ and $\mathbf{H}^{2}$ respectively. It is preferred that the mutual information between $\mathbf{H}^{1}$ and $\mathbf{H}^{2}$ is maximized, for computational reason its lower bound is usually used as the objective for contrastive learning. We use InfoNCE \cite{PCL} as our node-wise contrastive loss, that is,
\begin{equation}
\label{eq:node-level_loss}
	\mathcal{L}_{node} = \sum_{i=1}^N  -\log \frac{s(\mathbf{h}^1_i, \mathbf{h}^2_i)}{s(\mathbf{h}^1_i, \mathbf{h}^2_i)+ \sum_{j=1}^{N} s(\mathbf{h}^1_i, \mathbf{h}^2_j) } ,
\end{equation}
where $s(\mathbf{h}^1_i, \mathbf{h}^2_i) = \frac{\abs{\langle \mathbf{h}^1_i, \mathbf{h}^2_i \rangle}}{\|\mathbf{h}^1_i\|_2 \|\mathbf{h}^2_i\|_2}$ is the cosine similarity between two node representations, $\mathbf{h}^1_i$ and $\mathbf{h}^2_i$.

% We denote the node representation obtained by the GNN encoder as $\mathbf{H}=g(\mathbf{X}, \mathbf{A})$, with $h_i$
In addition to the node-wise contrastive learning, we also adopt prototypical contrastive learning \cite{PCL} to capture semantic information in the node representations, which can be interpreted as maximizing the mutual information between node representation and a set of estimated cluster prototypes $\{\bm{c}_1,...,\bm{c}_{{K}}\}$. Here $K$ is the number of cluster prototypes.
% $\bm{c}_k$ is the prototype featuring class $k$.
The loss function for prototypical contrastive learning is
\begin{equation}
\label{eq:proto_loss}
    \mathcal{L}_{proto} = -\frac{1}{N}\sum_{i=1}^N\log\frac{\exp(\mathbf{h}_i\cdot \mathbf{c}_{k}/\tau)}{\sum_{k=1}^{{K}} \exp(\mathbf{h}_i \cdot \mathbf{c}_k/\tau)}.
\end{equation}

BRGCL aims to improve the robustness of node representations by prototypical contrastive learning. Our key observation is that there exist a subset of nodes which are quite confident about their class/cluster labels because they are faraway from class/cluster boundaries. We propose an effective method to infer such confident subset of nodes. Because the BRGCL encoder is completely unsupervised, it does not have access to the ground truth label or ground truth class/cluster number. Therefore, our algorithm for selection of confident nodes is based on Bayesian non-parameter style inference, and the algorithm is termed Bayesian nonparametric Estimation of Confidence (BEC) to be introduced next.

\subsection{Bayesian nonparametric Estimation of Confidence (BEC)}
\label{section:BEC}
The key idea of Bayesian nonparametric Estimation of Confidence (BEC) is to estimate robust nodes by the confidence of nodes in their labels. Intuitively, nodes more confident in their labels are less likely to be adversely affected by noise. Because BRGCL is unsupervised, pseudo labels are used as the labels for such estimation.

We propose Bayesian nonparametric Prototype Learning (BPL) to infer the pseudo labels of nodes. BPL, as a Bayesian nonparametric algorithm, infers the cluster prototypes by the Dirichlet Process Mixture Model (DPMM) under the assumption that the distribution of node representations is a mixture of Gaussians. The Gaussians share the same fixed covariance matrix $\sigma \mathbf{I}$, and each Gaussian is used to model a cluster. The DPMM model is specified by
\begin{equation}
\begin{aligned}
G \sim  \text{DP} (G_0, \alpha),
\bm\phi_i  \sim  G,
\mathbf{h}_i  \sim  \mathcal{ N}(\bm\phi_i,\sigma  \mathbf{I}), \,\,\, i = 1, ..., N,
\end{aligned}
\label{eq:DPMM}
\end{equation}
where $G$ is a Gaussian distribution draw from the Dirichlet process $\mbox{DP}(G_0, \alpha)$, and $\alpha$ is the concentration parameter for $\mbox{DP}(G_0, \alpha)$. $\phi_i$ is the mean of the Gaussian sampled for generating the node representation $\mathbf{h}_i$. $G_0$ is the prior over means of the Gaussians. $G_0$ is set to a zero-mean Gaussian ${\mathcal N}(\bm{0},\rho \mathbf{I})$ for $\rho > 0$. A collapsed Gibbs sampler \cite{resnik2010gibbs} is used to infer the components of the GMM with the DPMM.
The Gibbs sampler iteratively samples pseudo labels for the nodes given the means of the Gaussian components, and samples the means of the Gaussian components given the pseudo labels of the nodes.
Following \cite{kulis2011revisiting}, such process is almost equivalent to K-means when $\sigma$, the variance of the Gaussians, goes to $0$. The almost zero variance eliminates the needs of estimating the variance $\sigma$, making the inference efficient.
% the process to sample pseudo labels becomes deterministic \cite{kulis2011revisiting}.

Let $\Tilde{K}$ denote the number of inferred prototypes at the  current iteration, the pseudo label $z_{i}$ of node $v_i$ is then calculated by
\begin{equation}
\label{eq:label_update}
\begin{aligned}
  z_{i} = \argmin_{k} \left\{ d_{ik} \right\},\, i=1,...,N, \,\,\,\,
  d_{ik} =
          \left\{\begin{array}{l l}
          \|\mathbf{h}_{i} - \mathbf{c}_{k}\|_2^2  & \,\, k = 1,..., \Tilde{K}, \\
          \xi  & \,\, k = \Tilde{K}+1,
        \end{array}\right.
%   z_{i} &=
%           \left\{\begin{array}{l l}
%           K+1  &\text{if~} \min_k d_{ik} > \beta\\
%           \argmin_{k}d_{ik} &\text{otherwise}
%         \end{array}\right.
\end{aligned}
\end{equation}
% where $\xi = 2\sigma \log(\frac{\alpha}{(1+\rho/\sigma)^{1/2}})$. In practice, we choose $\xi$ with cross-validation.
where the Euclidean distance $\left\{d_{ik}\right\}$ is used to determine the pseudo labels of the node representation $\mathbf{h}_i$. $\xi$ is the margin to initialize a new prototype.
% $\xi$ is actually related to the concentration parameter $\alpha$.
In practice, we choose the value of $\xi$ by performing cross-validation on each dataset with details in Section \ref{sec:imp_details} of the supplementary.
% $d_{ik}$ is to measure the Euclidean distance between the node representation $\mathbf{h}_i$ and prototype representation $\mathbf{c}_{k}$.
% With the formulation in Equation (\ref{eq:label_update}), a node will be assigned to the prototype modeled by the component corresponding to the closest mean of Gaussian, unless the squared Euclidean distance to the closest mean is greater than $\xi$.
% In this case, we initialize a component with the representation of this node.

After obtaining the pseudo labels of nodes by BPL with $K$ being the inferred number of prototypes, we estimate the confidence of the nodes based on their pseudo labels and the graph structure. We first select nodes confident in their labels, also referred to as confident nodes, by considering the label information from the neighborhood of each node specified by the adjacency matrix. Let $\bm{z}_i$ denote the one-hot pseudo label of node $v_i$ estimated by the Bayesian method. Label propagation~\cite{zhang2018link} is applied based on the adjacency matrix to get a soft pseudo label for each node.
Let $\mathbf{Z} \in \mathbb R^{N \times K}$ be the matrix of pseudo labels with $\mathbf{z}_i$ being the $i$-th row of $\mathbf{Z}$. The label progation runs the following update for $T$ steps,
\begin{equation}
\label{eq:lp}
            \mathbf{Z}^{(t+1)} = (1 - \alpha) \Tilde{\mathbf{A}} \mathbf{Z}^{(t)} + \alpha\mathbf{Z} ~~~t=1,...,T-1,
\end{equation}
where $T$ is the number of propagation steps, $\alpha$ is the teleport probability, which are set to the suggested values in \cite{zhang2018link}. Let $\Tilde{\mathbf{Z}} = \mathbf{Z}^{(T)}$ be the soft labels obtained by the label propagation with $\bm{\Tilde{z}_i}$ being the $i$-th row of $\Tilde{\mathbf{Z}}$. Following \cite{Han2018NIPS}, we use the cross-entropy between $\bm{z}_i$ and $\bm{\Tilde{z}_i}$, denoted by $\phi(\bm{z}_i,\bm{\Tilde{z}_i})$, to identify confident nodes.  Intuitively, smaller cross-entropy $\phi(\bm{z}_i,\bm{\Tilde{z}_i})$ of a node $v_i$ leads to a larger probability of the pseudo label, so node $v_i$ is more confident about its pseudo label $\bm{\Tilde{z}_i}$. As a result, we denote the set of confident nodes assigned to the $k$-th cluster as
\begin{equation}
	\mathcal{T}_k=\{\bm{h}_i\mid \phi(\bm{z}_i, \tilde{\bm{z}}_i)\textless \gamma_k\},
	\label{eq:confidence}
\end{equation}	
where $\gamma_k$ is a threshold for the $k$-th class. Figure~\ref{fig:aggregation} illustrates the cross-entropy values of all the nodes for the case that different levels of noisy are present in the input node attributes, where heat value indicates the corresponding cross-entropy value for every node. The confident nodes with less cross-entropy values, which are marked in more red, are far away from cluster boundaries, so that noise on these nodes are more unlikely to affect their classification/clustering labels. These confident nodes are the robust nodes leveraged by BRGCL to fight against noise.

The threshold $\gamma_k$ is dynamically set by
\begin{equation}
\label{eq:conf_thres}
    \gamma_k = 1 - \min\left\{\gamma_0, \gamma_0 \frac{t}{t_{\text{max}}}\right\},
\end{equation}
where $t$ is the current epoch number and $t_{\text{max}}$ is the number of epochs for training. In Co-teaching \cite{Han2018NIPS}, a similar threshold is used to select a ratio of data for training. However, due to the limited size of training data in graph domain, training with only a subset of nodes usually leads to degraded performance. For example, with $5\%$ of nodes labeled on Cora dataset, only $1\%$ of nodes will be used for training if the threshold is set to $20\%$ by Co-teaching. In contrast, BEC select confident nodes by a dynamic threshold on the confidence of nodes in their labels given the labels from their neighbors. The selected confident nodes are only used to obtain the robust prototype representations, and BRGCL is trained with such robust prototypes to obtain robust representations for all the nodes of the graph.

$\gamma$ is an annealing factor. In practice, the value of $\gamma_0$ is decided by cross-validation for each dataset, with details in Section \ref{sec:imp_details} of the supplementary. Previous methods such as \cite{PCL} estimate each prototype as the mean of node representations assigned to that prototype. After acquiring the confident nodes $\left\{\mathcal{T}_k\right\}_{k=1}^{K}$, the prototype representations are updated by $\bm{c}_k = \frac{1}{|\mathcal{T}_k|} \sum_{\bm{h}_i\in \mathcal{T}_k} \bm{h}_i$ for each $k \in [K]$. With the updated cluster prototypes $\left\{\bm{c}_k\right\}_{k=1}^K$ in the prototypical contrastive learning loss (\ref{eq:proto_loss}), we train the encoder $g(\cdot)$ with the following overall loss function,
\begin{equation}
\label{eq:gaussian_lower}
    \mathcal{L}_{rep} = \mathcal{L}_{node} + \mathcal{L}_{proto}.
\end{equation}

Training BRGCL with the loss function $\mathcal{L}_{rep}$ does not require any information about the ground truth labels. We summarize the training algorithm for the BRGCL encoder in Algorithm~\ref{Algorithm-BRGCL}. It is noted that confident nodes and robust prototypes are estimated at each epoch by BEC.

\begin{algorithm}[tbp]\caption{Training algorithm of BRGCL encoder}\label{Algorithm-BRGCL}
{
\small
\begin{algorithmic}[1]
\REQUIRE The input attribute matrix $\mathbf{X}$, adjacency matrix $\mathbf{A}$, and the training epochs $t_{\text{max}}$.
\ENSURE The parameter of BRGCL encoder $g$.
\STATE Initialize the parameter of BRGCL encoder $g$
\FOR{$t\leftarrow 1$ to $t_{\text{max}}$}
\STATE Calculate node representations by $\mathbf{H} = g(\mathbf{X}, \mathbf{A})$
\STATE Generate augmented views $G^{1} = (\mathbf{X}^{1}, \mathbf{A}^{1})$ and $G^{2} = (\mathbf{X}^{2}, \mathbf{A}^{2})$
\STATE Calculate node representations of augmented views by $\mathbf{H}^{1}=g(\mathbf{X}^{1}, \mathbf{A}^{1})$ and $\mathbf{H}^{2}=g(\mathbf{X}^{2}, \mathbf{A}^{2})$
% by Eq.(\ref{eq:aug_rep})
\STATE Calculate loss $\mathcal{L}_{node}$ by Eq. (\ref{eq:node-level_loss})
\STATE Obtain the pseudo labels of all the nodes $\mathbf{Z}$ and the number of inferred prototypes $K$ by Eq. (\ref{eq:label_update})
\STATE Obtain soft labels of nodes $\mathbf{\Tilde{Z}}$ by label propagation in Eq. (\ref{eq:lp})
\STATE Update the confidence thresholds $\{\gamma_k\}_{k=1}^{K}$ by Eq. (\ref{eq:conf_thres})
\STATE Estimate the sets of confident nodes $\{\mathcal{T}_k\}_{k=1}^{K}$ by Eq. (\ref{eq:confidence})
\STATE Update confident prototype representations by $\bm{c}_k = \frac{1}{|\mathcal{T}_k|} \sum_{\bm{h}_i\in \mathcal{T}_k} \bm{h}_i$ for all $k \in [K]$
%\STATE Calculate prototypical loss $\mathcal{L}_{proto}$ by Eq.(\ref{eq:gaussian_lower})
%\STATE Calculate overall loss $\mathcal{L}_{rep} = \mathcal{L}_{node} + \mathcal{L}_{proto}$
\STATE Update the parameter of BRGCL encoder $g$ using the loss $\mathcal{L}_{rep} $ in Eq. (\ref{eq:gaussian_lower})
\ENDFOR
\STATE \textbf{return} The BRGCL encoder $g$
\end{algorithmic}
}
\end{algorithm}

\subsection{Decoupled Training}
The typical pipeline for semi-supervised node classification is to jointly train the classifier and the encoder. However, the noise in the training data would degrade the performance of the classifier.
%In previous node classification models, the classifier weights are usually trained jointly with the model parameters for extracting the node representation by minimizing the cross-entropy loss between the ground truth and predictions of labeled nodes. This is also a typical pipeline for node classification on inputs with noise. However, in this procedure, especially with a high noise ratio in the training labels, the training for the classifier would greatly degrade the node representation learning.
To alleviate this issue, we decouple the representation learning for the nodes from the classification of nodes to mitigate the effect of noise, which consists of two steps. In the first step, the BRGCL encoder $g(\cdot)$ is trained by Algorithm~\ref{Algorithm-BRGCL}. In the second step, with the node representation $\mathbf{H}$ from the trained BRGCL encoder, the classifier $f(\cdot)$ is trained by optimizing the loss function $\LL_{\textup{cls}}$. In Section \ref{sec:decoupled} of the supplementary, we show the advantage of such decoupled learning pipeline over the conventional joint training of encoder and classifier.

\section{Experiments}
In this section, we evaluate the performance of BRGCL on five public benchmarks including the challenging large-scale dataset ogbn-arxiv, with details deferred to Section~\ref{sec:data} of the supplementary. For semi-supervised node classification, the performance of BRGCL is evaluated with noisy label or noisy input node attributes. For node clustering, only noisy input node attributes are considered because there are no ground truth labels given for clustering purpose. The implementation details about node classification is deferred to Section~\ref{sec:node-classification-details} of the supplementary.
\subsection{Experimental Settings}
Due to the fact that most public benchmark graph datasets do not come with corrupted labels or attribute noise, we manually inject noise into public datasets to evaluate our algorithm. We follow the commonly used label noise generation methods from the existing work \cite{han2020survey} to inject label noise. We generate noisy labels over all classes according to a noise transition matrix $Q^{K\times K}$, where $Q_{ij}$ is the probability of nodes from class $i$ being flipped to class $j$. We consider two types of noise: (1) \textbf{Symmetric}, where nodes from each class can be flipped to other classes with a uniform random probability, s.t. $Q_{ij} = Q_{ji}$; (2) \textbf{Asymmetric}, where mislabeling only occurs between similar classes. The percentage of nodes with flipped labels is defined as the label noise level in our experiments. To evaluate the performance of our method with attribute noise, we randomly shuffle a certain percent of input attributes for each node following \cite{ding2022data}. The percentage of shuffled attributes is defined as the attribute noise level in our experiments.

In our experiments for node classification, noise is only applied to the training set. Both validation and test sets are kept clean. All models in our experiments are trained on the corrupted training data and tested on the clean test data.

In our experiments, we compare PRGCL against GCN~\cite{kipf2017semi}, GCE~\cite{zhang2018generalized}, S$^2$GC~\cite{zhu2020simple}, UnionNet~\cite{li2021unified}, and NRGNN~\cite{dai2021nrgnn}. The training for different baselines are categorized into two setups: (1) \textbf{Unsupervised Setup}, where the training of the encoder does not use the ground truth label information. The node representations obtained by the encoder are then used for downstream tasks, which are node classification and node clustering; (2) \textbf{Supervised Setup}, where the training of the encoder uses the ground truth label information. Our proposed BRGCL follows the unsupervised setup in all our experiments, and every baseline follows its corresponding setup by its nature.

\subsection{Evaluation Results}
\label{sec:results}
\begin{figure}[tbp]
    \centering
    \subfigure[Cora]{\includegraphics[width=0.3\textwidth]{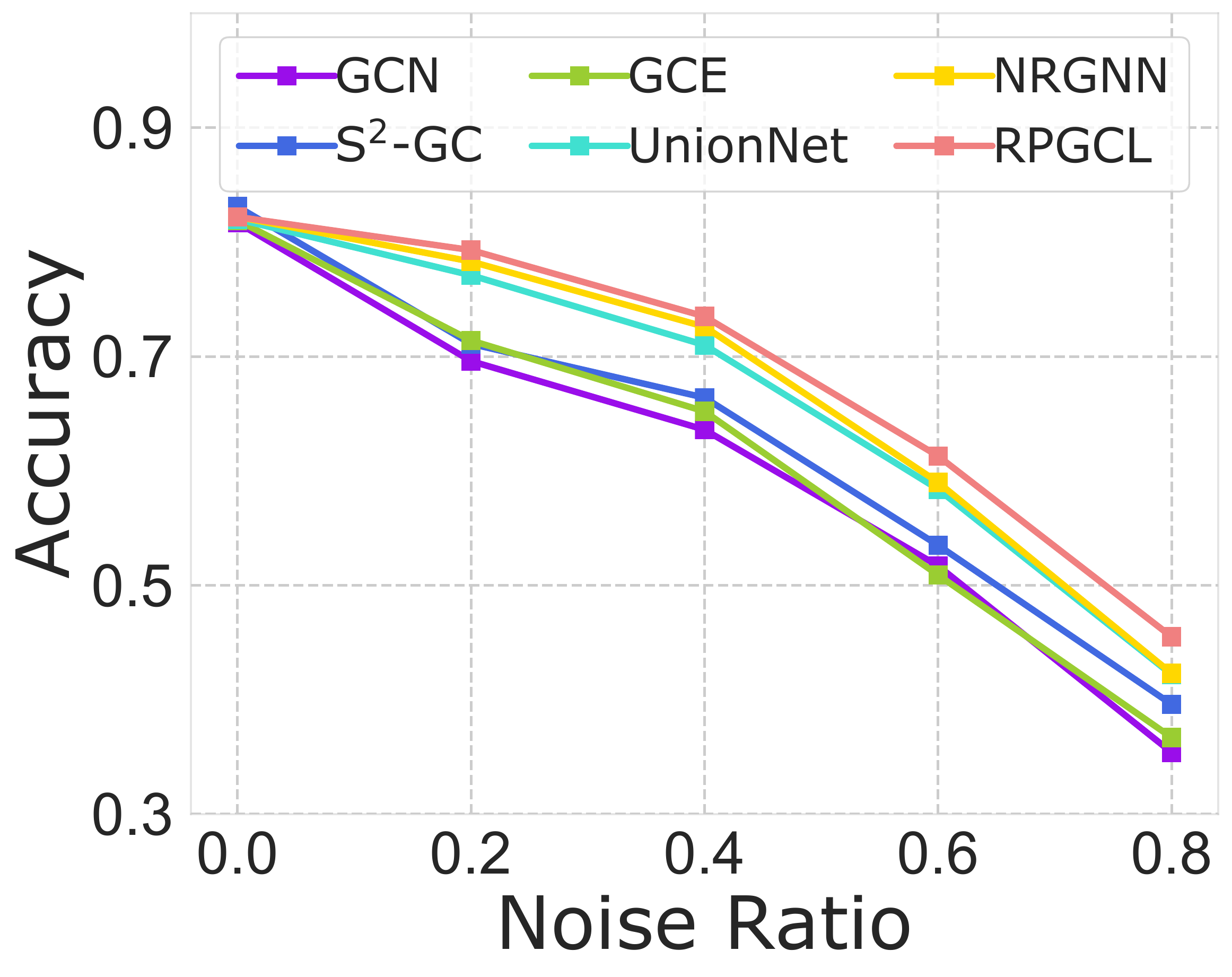}}
    \subfigure[Citeseer]{\includegraphics[width=0.3\textwidth]{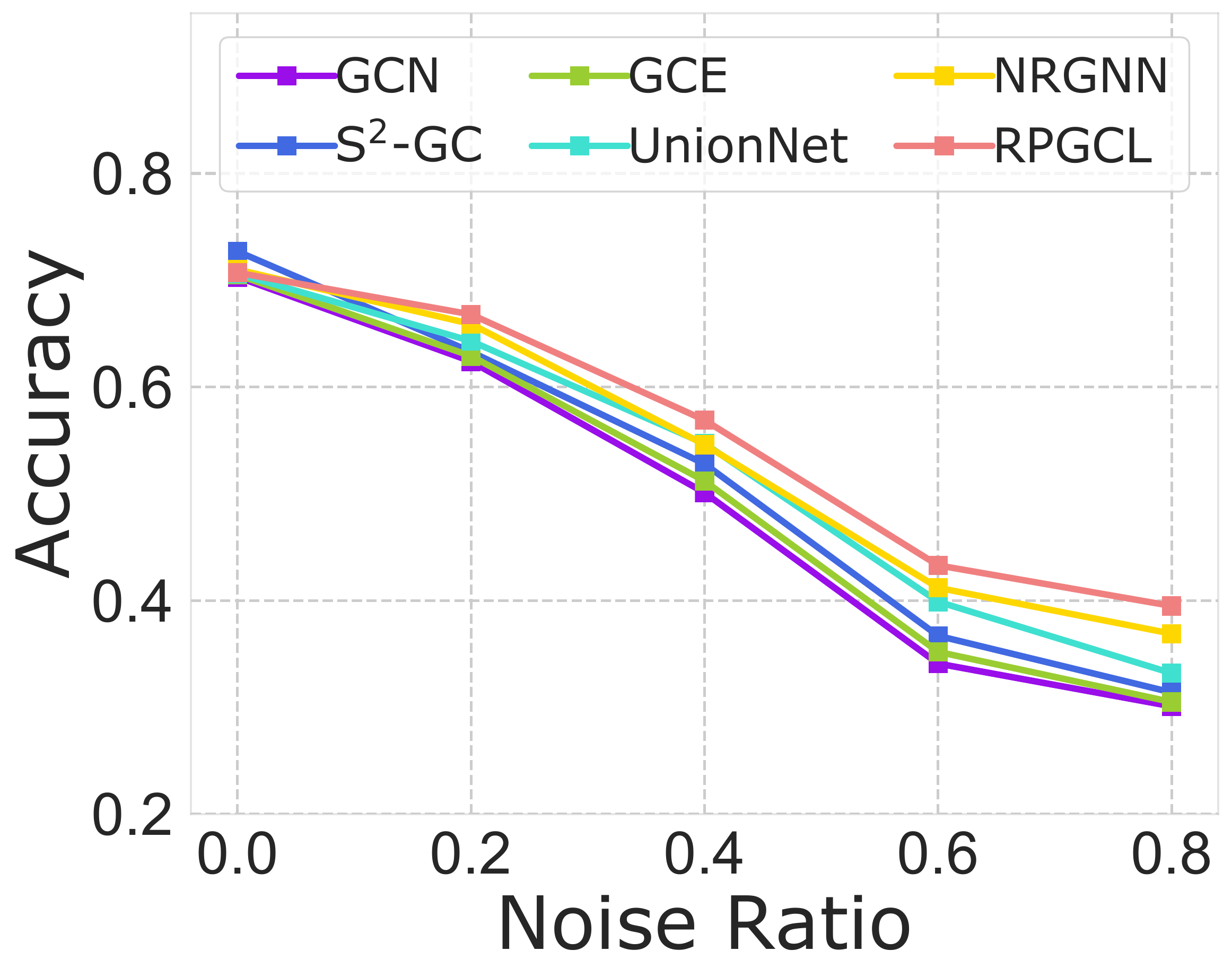}}
    \subfigure[Pubmed]{\includegraphics[width=0.3\textwidth]{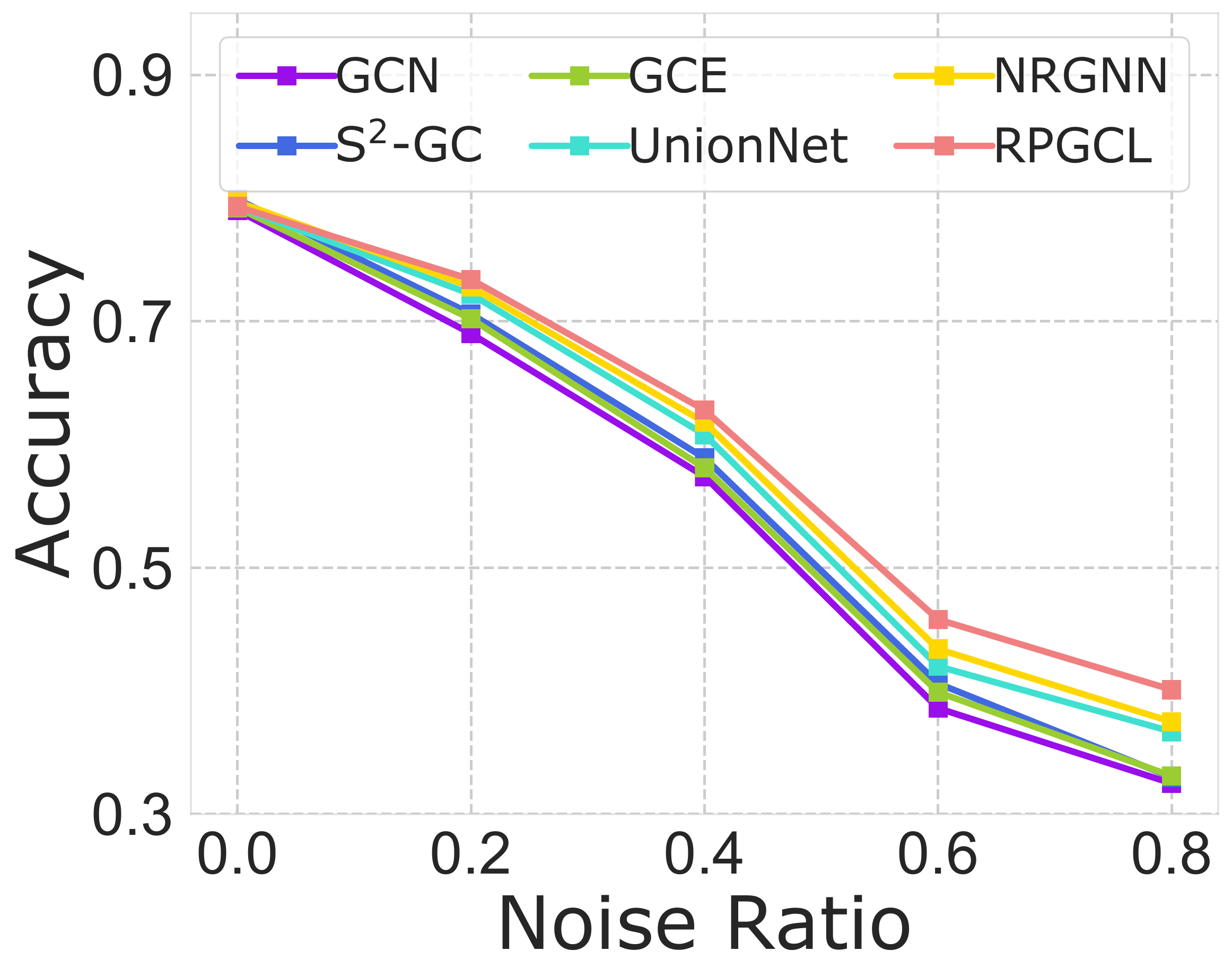}}
    \subfigure[Coauthor CS]{\includegraphics[width=0.3\textwidth]{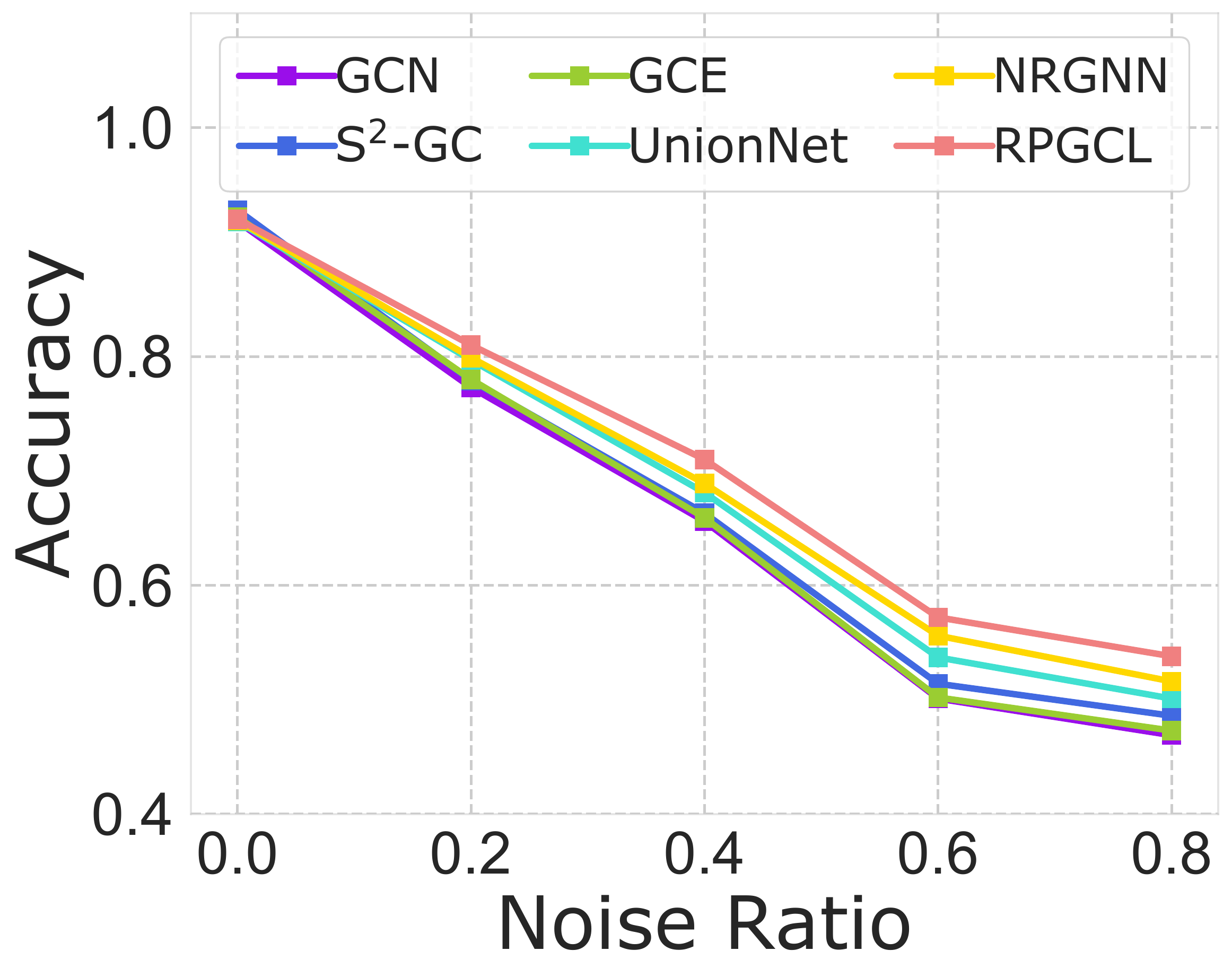}}
    \subfigure[ogbn-arxiv]{\includegraphics[width=0.3\textwidth]{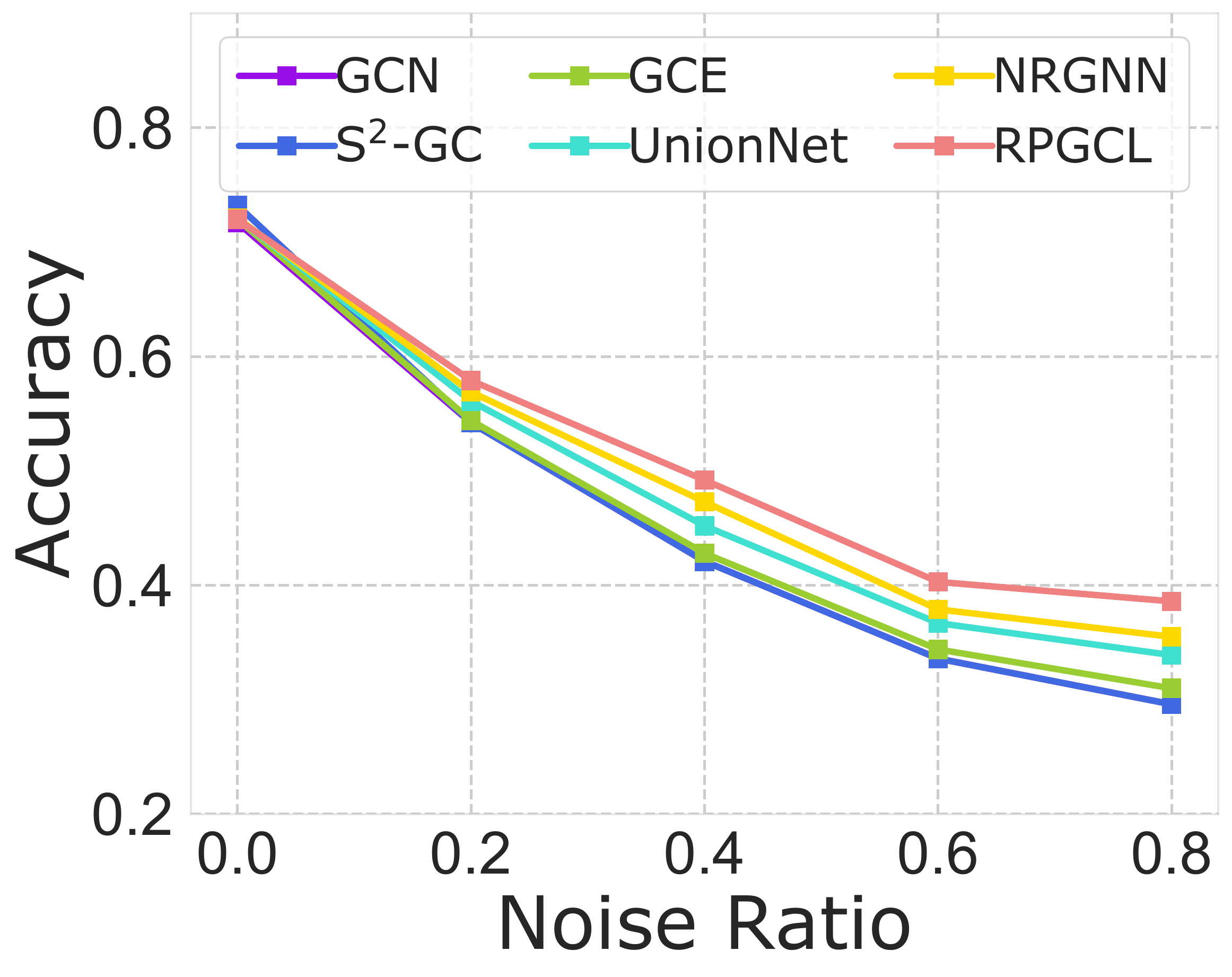}}
    \vspace{-0.1cm}
    \caption{Performance comparisons on semi-supervised node classification with different levels of symmetric label noise.}
    \label{fig:label_noise_result}
\end{figure}
\textbf{Semi-supervised Node Classification with Label Noise.}
We compare BRGCL against competing methods for semi-supervised node classification on input with two types of label noise. To show the robustness of BRGCL against label noise, we perform the experiments on graphs injected different levels of label noise. The classification follows the widely used semi-supervised setting \cite{kipf2017semi}. Note the labels are only used for the training of the classifier. The BRGCL encoder generates node representations, and the classifier for node classification is trained on these node representations.

In our experiment, a two-layer MLP whose hidden dimension is $128$ is used as the classifier. The results of different methods with respect to different symmetric label noise levels are illustrated in Figure~\ref{fig:label_noise_result}. We report the means of the accuracy of $20$ runs and the standard deviation for all the baselines in Section~\ref{sec:detail_resutls} of the supplementary. The results for asymmetric label noise levels are in Section \ref{sec:asy_label} of the supplementary. It is observed from the results that BRGCL outperforms all the baselines including the methods using ground truth labels to train their encoders. By selecting confident nodes and computing robust prototypes using BEC, BRGCL outperforms all the baselines by an even larger margin with a larger label noise level.

\begin{figure}[tbp]
    \centering
    \subfigure[Cora]{\includegraphics[width=0.3\textwidth]{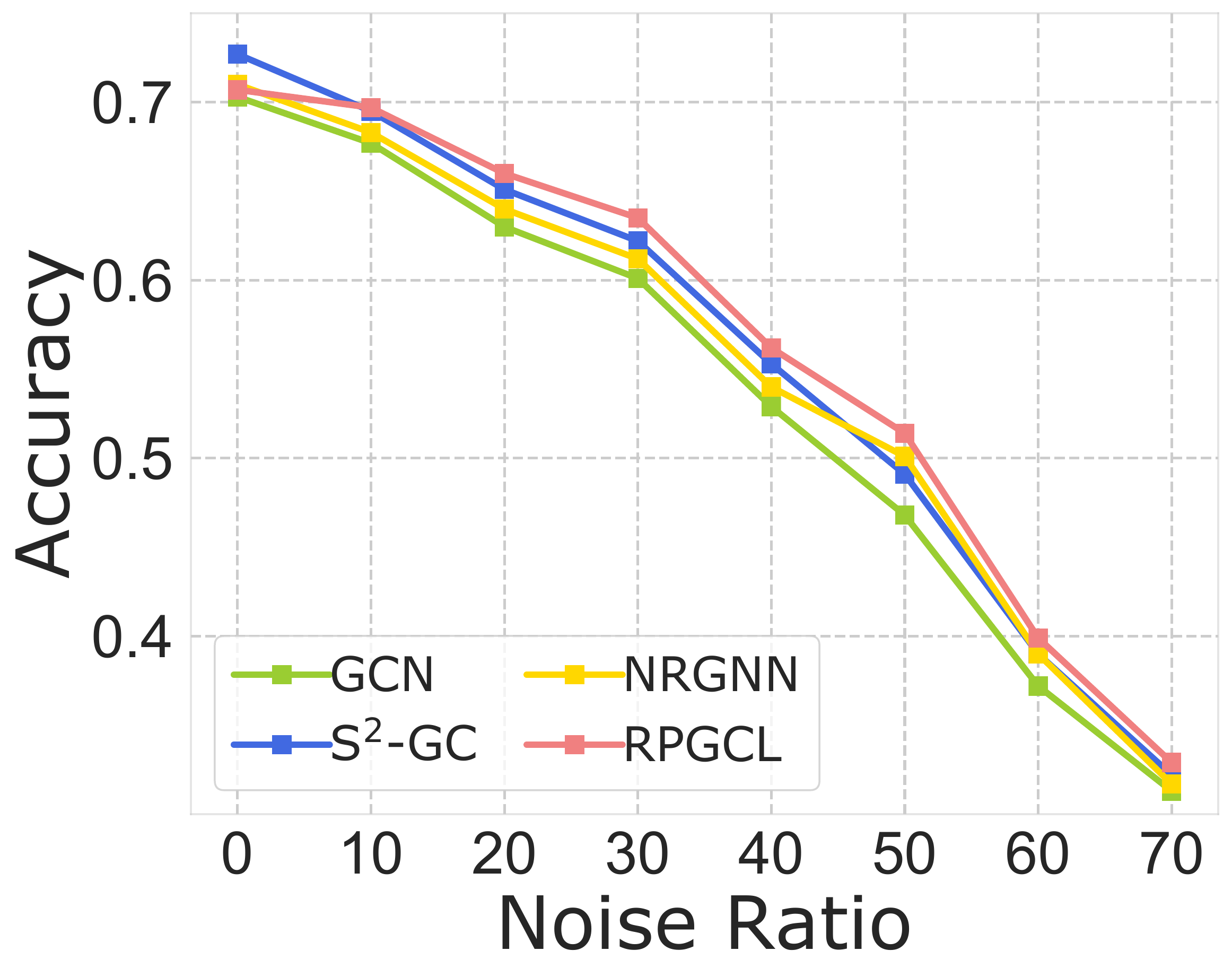}}
    \subfigure[Citeseer]{\includegraphics[width=0.3\textwidth]{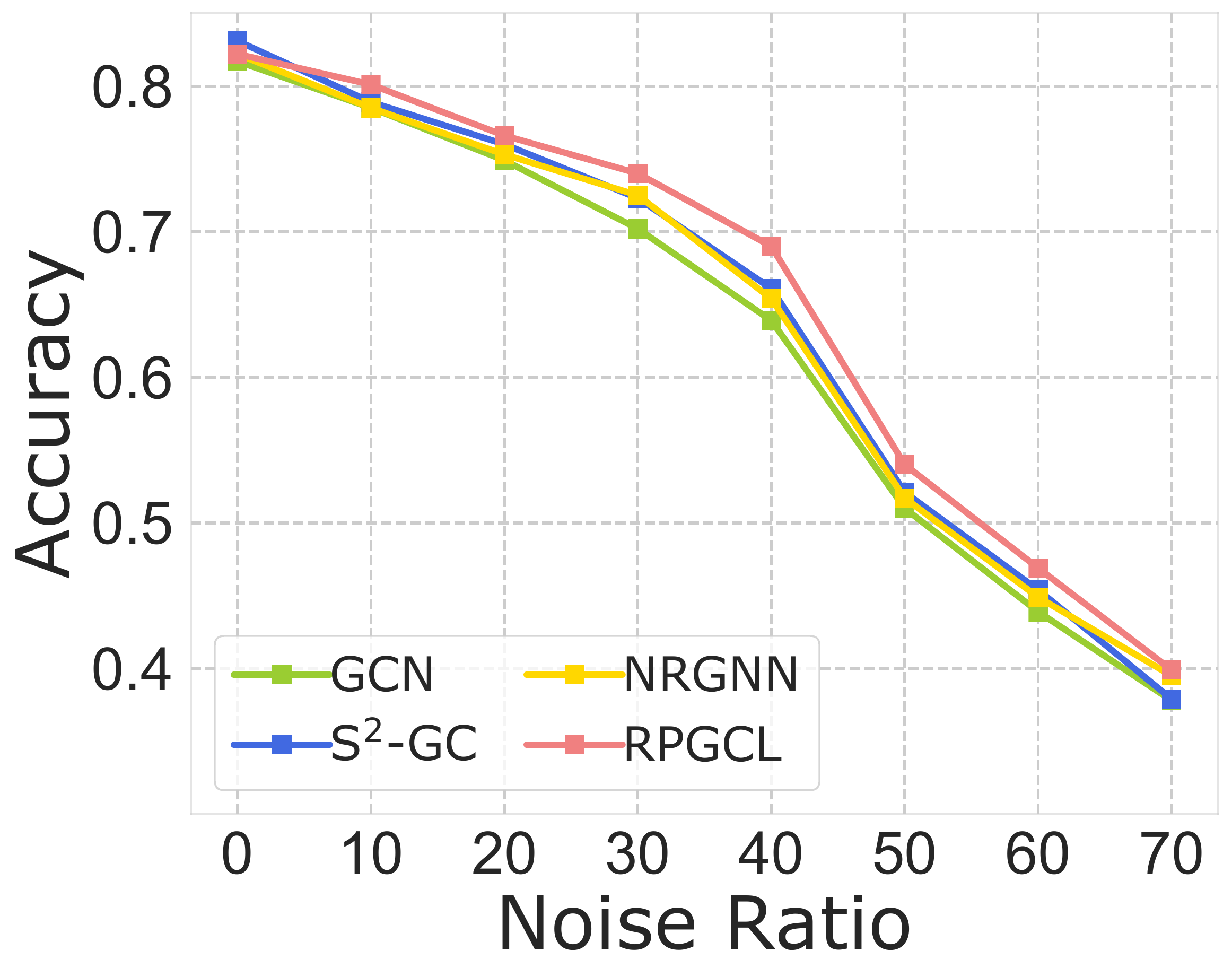}}
    \subfigure[Pubmed]{\includegraphics[width=0.3\textwidth]{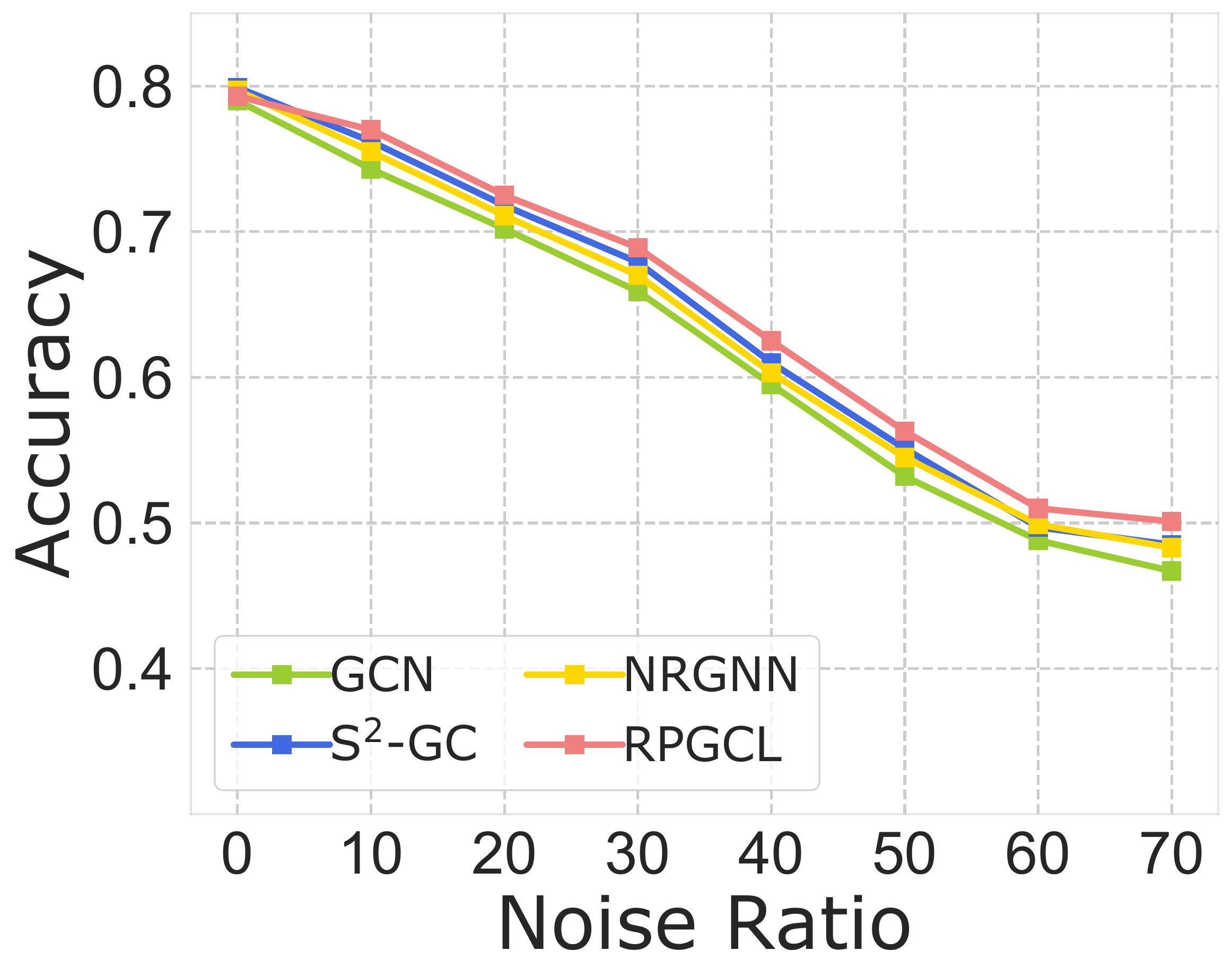}}
    \subfigure[Coauthor CS]{\includegraphics[width=0.3\textwidth]{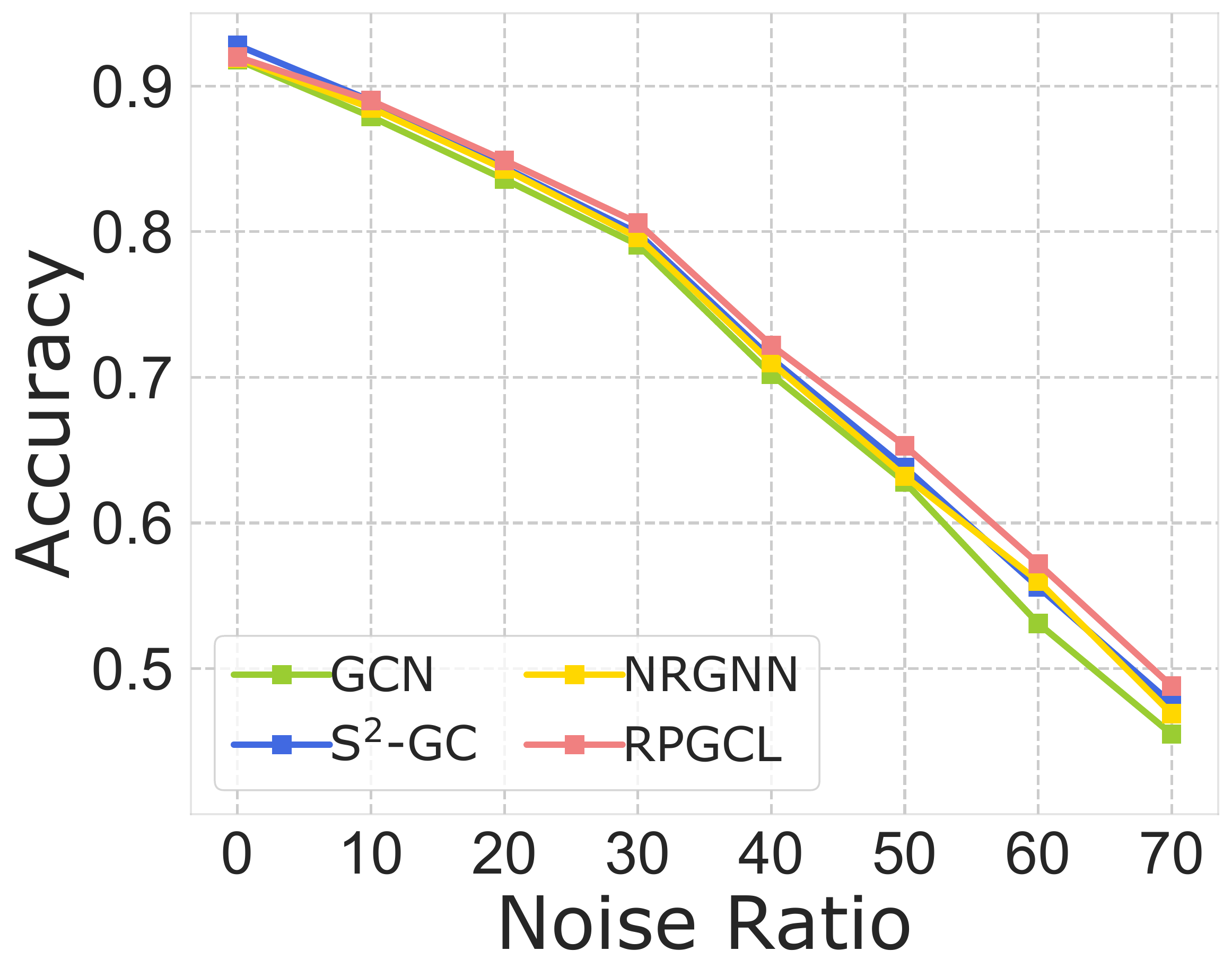}}
    \subfigure[ogbn-arxiv]{\includegraphics[width=0.3\textwidth]{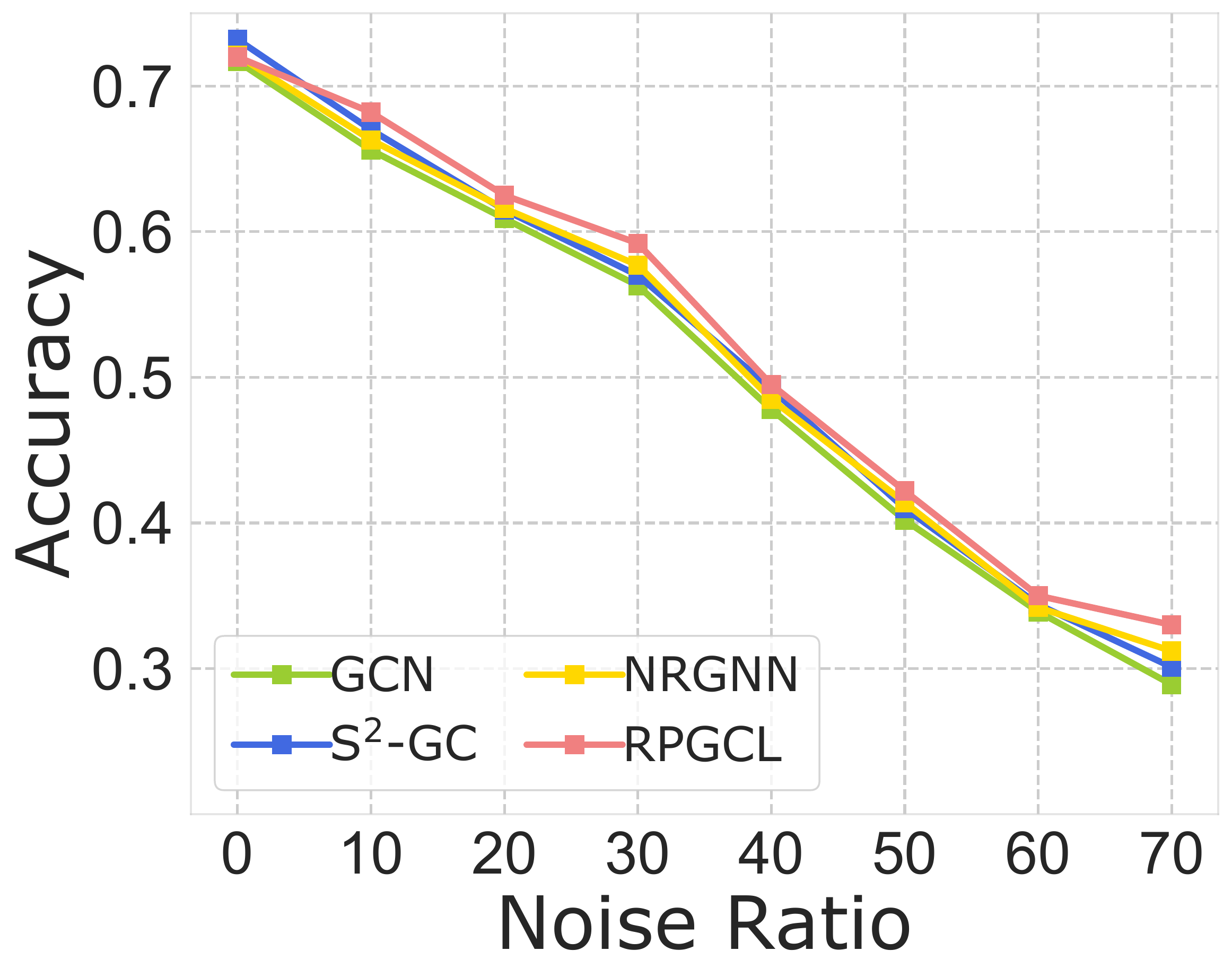}}
    \caption{Performance comparisons on semi-supervised node classification with different level feature noise.}
    \label{fig:feature_noise_result}
\end{figure}

\textbf{Semi-supervised Node Classification with Attribute Noise.}
We compare BRGCL with baselines for noisy input attributes with attribute noise level ranging from $0\%$ to $70\%$ with a step of $10\%$. GCE and UnionNET from semi-supervised node classification are excluded from the comparison as they are specifically designed for label noise. The results are illustrated in Figure~\ref{fig:feature_noise_result}, which clearly shows that BRGCL is more robust to attribute noise compared to all the baselines for different noise levels. We include detailed experimental results with standard deviation of $20$ runs in Section \ref{sec:detail_resutls} of the supplementary.

\textbf{Node Clustering}
To further evaluate the robustness of node representation learned by BRGCL, we perform experiments on node clustering. We follow the same evaluation protocol as that in \cite{hassani2020contrastive}. K-means is applied on the learned node representations to obtain clustering results. We use accuracy (ACC), normalized mutual information (NMI), and adjusted rand index (ARI) as the performance metrics for clustering. We report the averaged clustering results over $10$ times of execution in Table~\ref{tab:clustering}. We further report node clustering results with noisy attributes in Section~\ref{sec:clustering_noise} of the supplementary.
\begin{table*}[htbp]
\centering
\scalebox{0.85}{
\begin{tabular}{cccccccccc}
\hline
\multirow{2}{*}{Methods} & \multicolumn{3}{c}{Cora} & \multicolumn{3}{c}{Citeseer} & \multicolumn{3}{c}{Pubmed} \\ \cline{2-10}
                         & ACC    & NMI    & ARI    & ACC      & NMI     & ARI     & ACC     & NMI     & ARI    \\ \hline
\multicolumn{10}{c}{Supervised}                                                                                 \\ \hline
GCN                      & 68.3   & 52.3   & 50.9   & 68.8     & 41.9    & 43.1    & 69.1    & 31.2    & 31.7   \\
S$^2$GC                  & 69.6   & 54.7   & 51.8   & 69.1     & 42.8    & 43.5    & 70.1    & 33.2    & 32.2   \\
NRGNN                    & 72.1   & 55.6   & 52.1   & 69.3     & 43.6    & 44.7    & 69.9    & 34.2    & 32.8   \\ \hline
\multicolumn{10}{c}{Unsupervised}                                                                               \\ \hline
K-means                  & 49.2   & 32.1   & 22.9   & 54.0     & 30.5    & 27.8    & 59.5    & 31.5    & 28.1   \\
GAE                      & 59.6   & 42.9   & 34.7   & 40.8     & 17.6    & 12.4    & 67.2    & 27.7    & 27.9   \\
ARGA                     & 64.0   & 44.9   & 35.2   & 57.3     & 35.0    & 34.1    & 66.8    & 30.5    & 29.5   \\
ARVGA                    & 64.0   & 45.0   & 37.4   & 54.4     & 26.1    & 24.5    & 69.0    & 29.0    & 30.6   \\
GALA                     & 74.5   & 57.6   & 53.1   & 69.3     & 44.1    & 44.6    & 69.3    & 32.7    & 32.1   \\
MVGRL                    & 74.8   & 57.8   & 53.0   & 69.6     & 44.7    & 45.2    & 69.6    & 33.9    & 32.5   \\
BRGCL                     & \textbf{75.2}   & \textbf{58.3}   & \textbf{53.4}   & \textbf{70.1}     & \textbf{45.3}    & \textbf{46.2}    & \textbf{70.1}    & \textbf{35.1}    & \textbf{33.4}   \\ \hline
\end{tabular}
}
\vspace{-0.15cm}
\caption{Node clustering performance comparison on benchmark datasets. The three methods under the category ``Supervised'' use ground truth labels to train their encoders, while other methods under the category ``Unsupervised'' do not use the ground truth labels to train their encoders.}
\label{tab:clustering}
\end{table*}

\textbf{Comparison to Existing Sample Selection.} We also compare our BRGCL to the representatives sample selection methods for node classification, including Co-teaching \cite{Han2018NIPS}, in Section~\ref{sec:sample_selection} of the supplementary. It is observed that BRGCL outperforms these competing methods by a noticeable margin.

% \textbf{(1) label every baseline using ground truth labels or not; use plot to show the numbers (incremental of $10$ in noise level), put table of numbers to the supplementary; add the baseline S$^2$GC double check the literature of GCL with labels or no labels at all (unsupervised) (2) Visualize the heatmap of confidence scores. When having noisy feature results, such visualization should be with respect to the noise level}
%\vspace{-.2in}
\subsection{Confidence Score Visualization}
\label{sec:ablation}
\begin{figure}[h]
    \centering
    \subfigure[noise level = 0]{\includegraphics[width=0.275\textwidth]{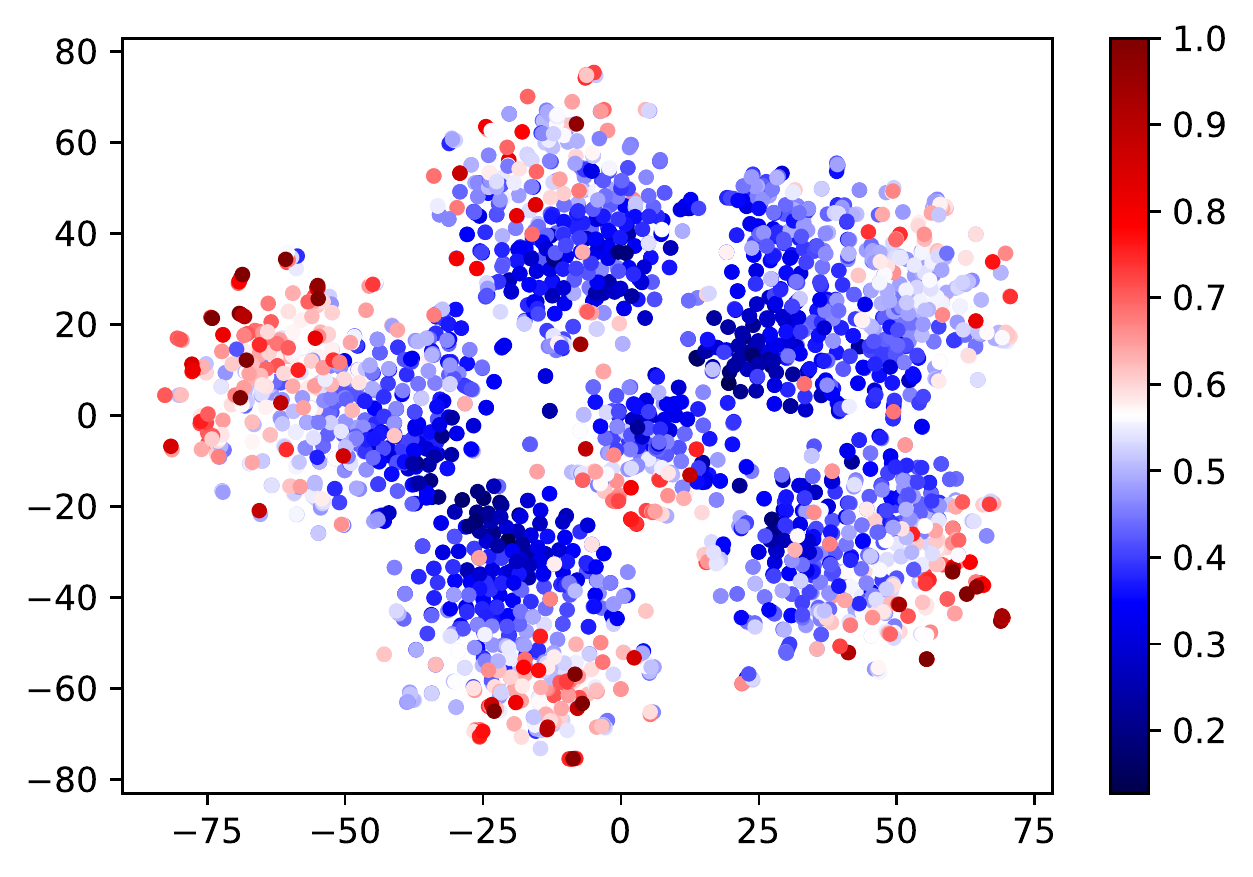}}
    \subfigure[noise level = 10]{\includegraphics[width=0.275\textwidth]{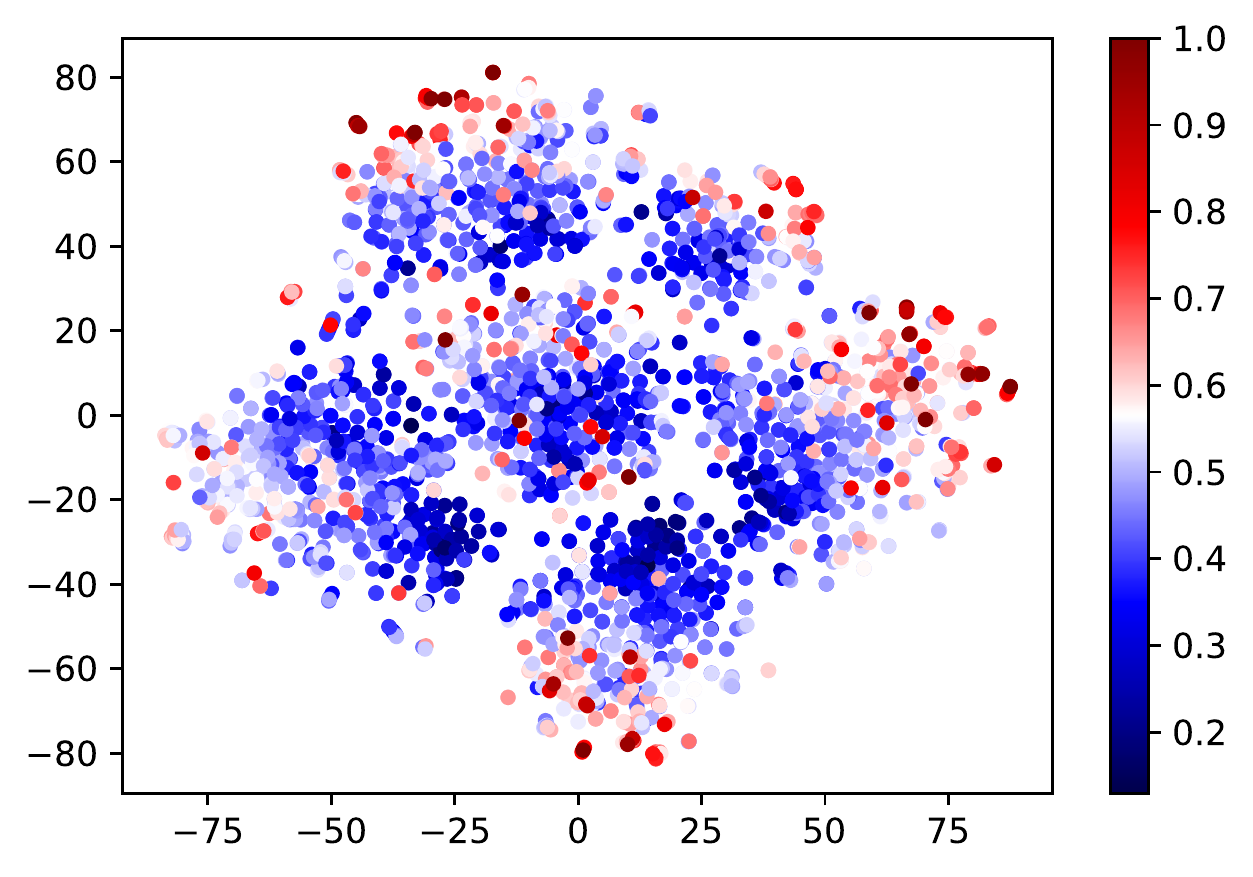}}
    \subfigure[noise level = 20]{\includegraphics[width=0.275\textwidth]{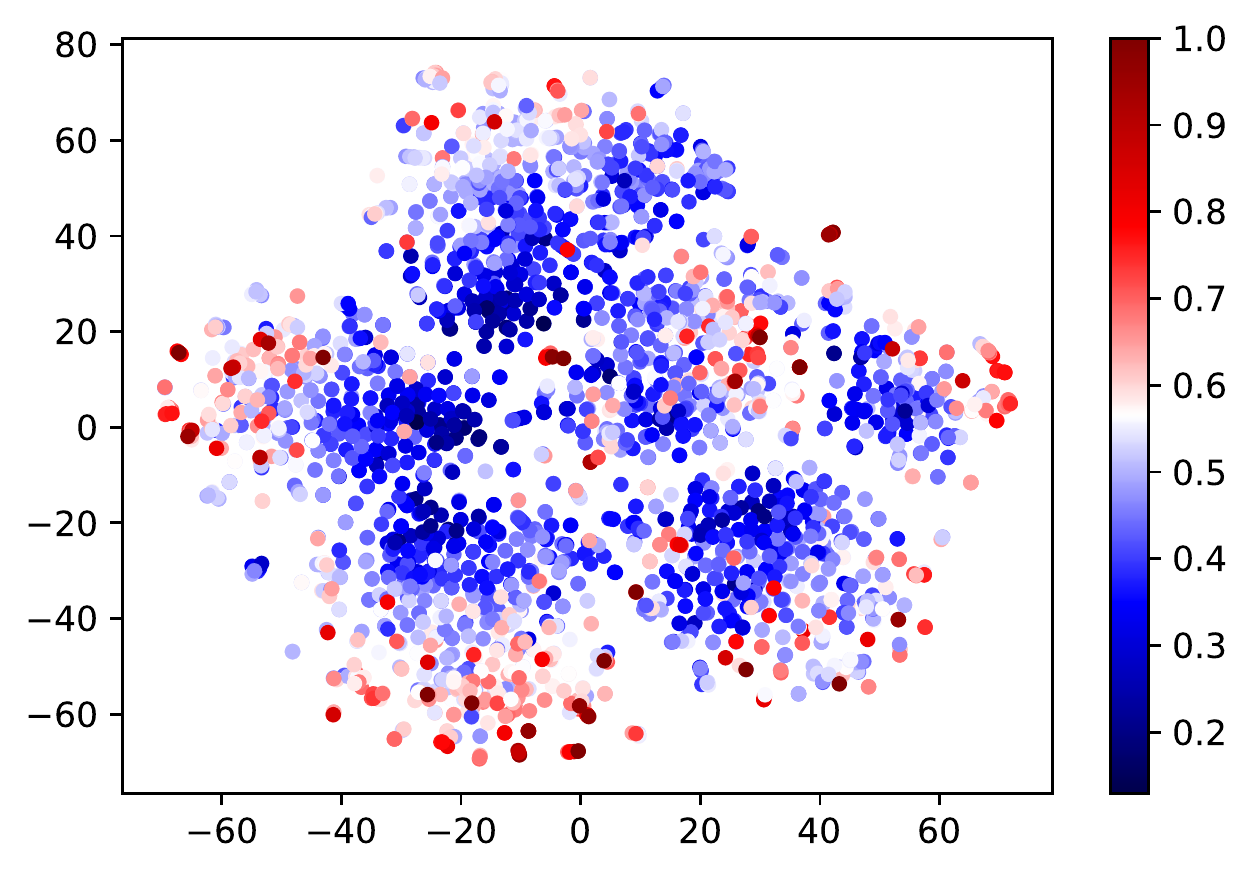}}
    \subfigure[noise level = 30]{\includegraphics[width=0.275\textwidth]{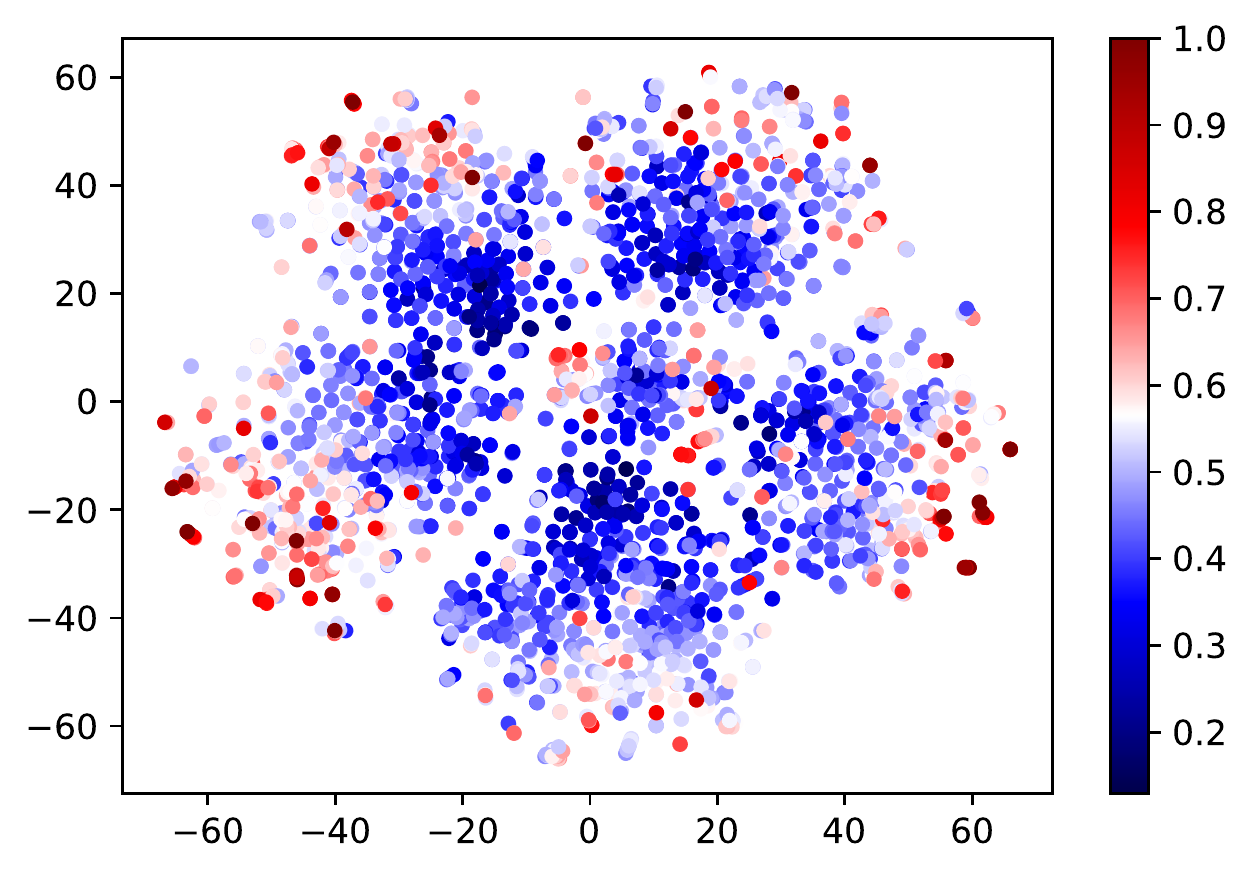}}
    \subfigure[noise level = 40]{\includegraphics[width=0.275\textwidth]{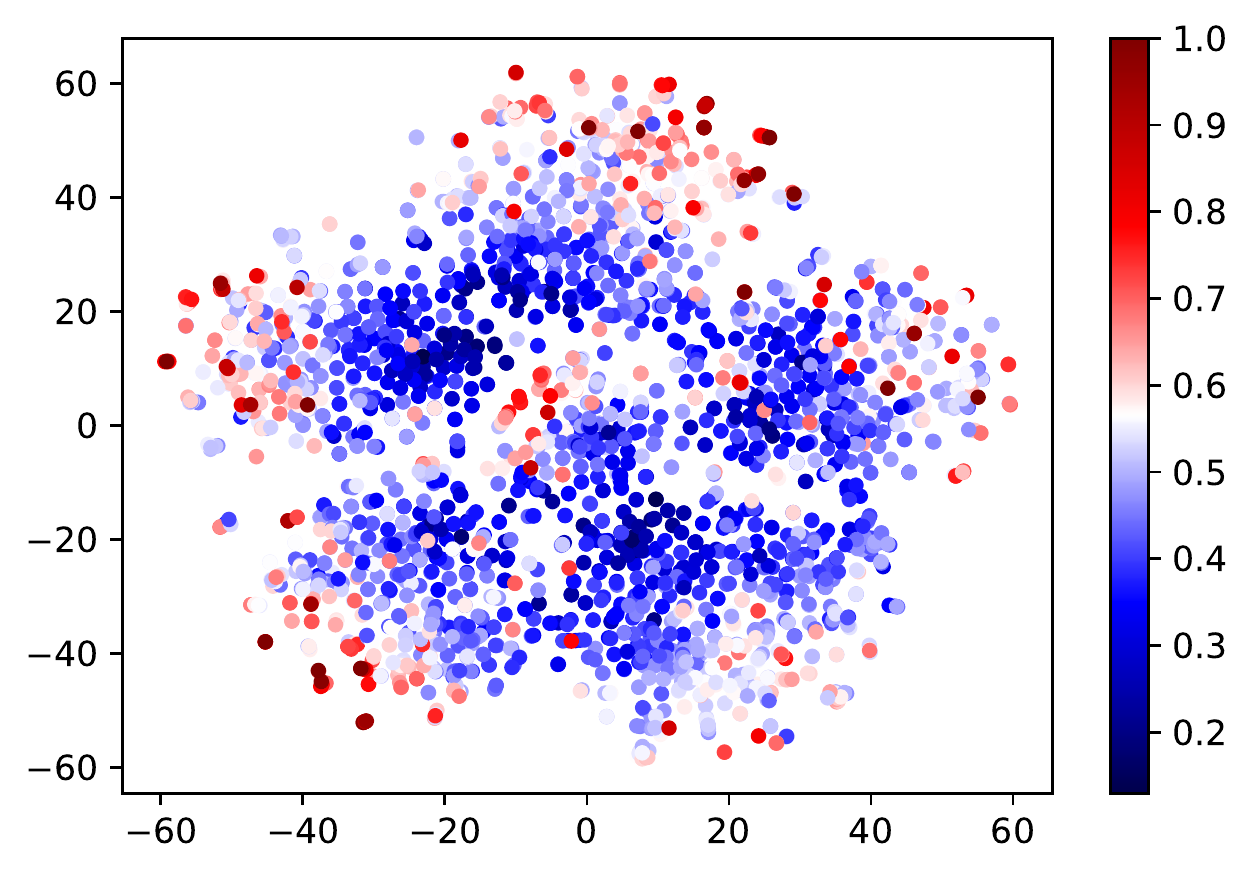}}
    \subfigure[noise level = 50]{\includegraphics[width=0.275\textwidth]{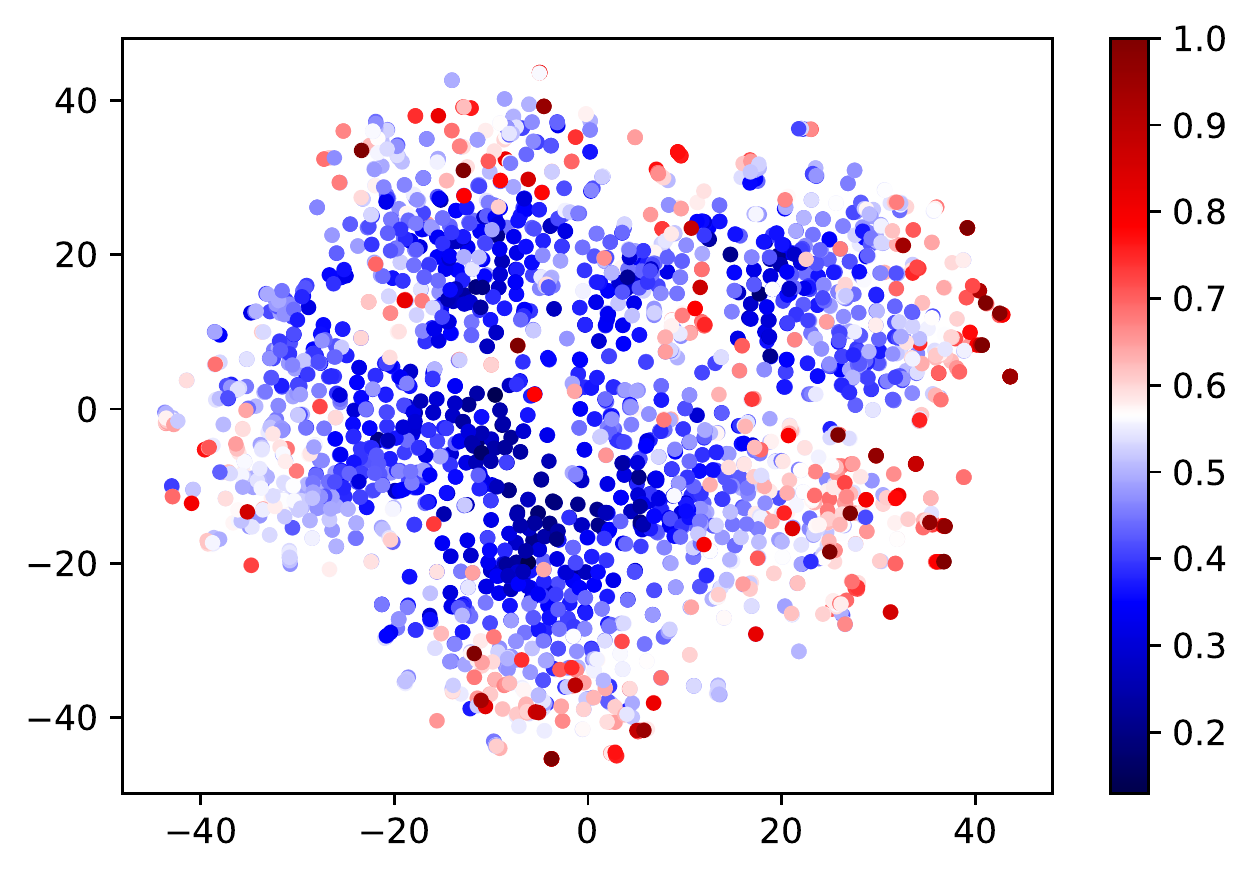}}
    \caption{Visualization of confident nodes with different levels of attribute noise for semi-supervised node classification.}
    \label{fig:aggregation}
\end{figure}

\vspace{-.1in}
We visualize the confident nodes selected by BEC in the embedding space of the learned node representations in Figure~\ref{fig:aggregation}. The node representations are visualized by t-SNE figure. Each mark in t-SNE represents the representation of a node, and the color of the mark denotes the confidence of that node. The results are shown for different levels of attribute noise. It can be observed from Figure~\ref{fig:aggregation} that confident nodes, which are redder in Figure~\ref{fig:aggregation}, are well separated in the embedding space. With a higher level of attribute noise, the bluer nodes from different clusters blended around the cluster boundaries. In contrast, the redder nodes are still well separated and far away from cluster boundaries, which leads to more robustness and better performance in downstream tasks.

\vspace{-.2in}
\section{Conclusions}
% \vspace{-.1in}
In this paper, we propose a novel node representation learning method termed Bayesian Robust Graph Contrastive Learning (BRGCL) that aims to improve the robustness of node representations by a novel Bayesian non-parametric algorithm,  Bayesian nonparametric Estimation
of Confidence (BEC). We evaluate the performance of BRGCL with comparison to competing baselines on semi-supervised node classification and node clustering, where graph data are corrupted with noise in either the labels for the node attributes.  Experimental results demonstrate that BRGCL generates more robust node representations with better performance than the current state-of-the-art node representation learning methods.

\newpage

{\small
\bibliographystyle{unsrt}
\bibliography{reference}
}

\newpage
\appendix
\section{Implementation Details}
\label{sec:imp_details}
\subsection{Datasets}
\label{sec:data}
We evaluate BRGCL on five public benchmarks that are widely used for node representation learning, namely Cora, Citeseer, Pubmed \cite{sen_2008_aimag}, Coauthor CS, and ogbn-arxiv \cite{hu2020open}. Cora, Citeseer and Pubmed are three most widely used citation networks.
Coauthor CS is co-authorship graph.
The ogan-arxiv is a directed citation graph.
We summarize the statistics of the datasets in Table~\ref{tab:dataset}.
Among the five benchmarks, ogbn-arxiv is known for its larger scale, and is more challenging to deal with. For all our experiments, we follow the default separation of training, validation, and test sets on each benchmark.
% The validation and test sets are kept the same as the cited papers to keep consistency. As for the training set, we randomly sample 5\% nodes for Cora and Citeseer. For large datasets, such as Pubmed and DBLP, we sample 1\% nodes to compose the training set. All the training set has no overlap with validation and test sets.
\begin{table}[h]
\small
	\centering
	\caption{The statistics of the datasets.}
	\vspace{-0.2cm}
	\begin{tabular}{@{}lcccc@{}}
		\toprule
		\textbf{Dataset} &  \textbf{Nodes} & \textbf{Edges} & \textbf{Features} & \textbf{Classes} \\ \midrule
		\textbf{Cora}              & 2,708    & 5,429        & 1,433            & 7                \\
		\textbf{CiteSeer}         & 3,327    & 4,732        & 3,703            & 6                \\
		\textbf{PubMed}           & 19,717   & 44,338      & 500               & 3                \\
		\textbf{Coauthor CS}     & 18,333     & 81,894       & 6,805              & 15            \\
		\textbf{ogbn-arxiv}     & 169,343     &  1,166,243       & 128     & 40            \\
 \bottomrule
	\end{tabular}
		
	\label{tab:dataset}
\end{table}

\subsection{More Details about Node Classification}
\label{sec:node-classification-details}
The robust node representations are used to perform node classification and node clustering mentioned in Section~\ref{sec:setup} of the main paper. More details about node classification are introduced in this subsection. As the connected neighbors in a graph usually show similar semantic information, we generate soft labels of nodes via label propagation on the graph to take the advantage of the information from the neighborhood. The classifier for node classification is trained with soft labels instead of hard labels.

First, we define the one-hot hard label matrix $\mathbf{Y} \in \mathbb{R}^{N \times K}$, where $\mathbf{Y}_{ij}=1$ if and only if node $v_i$ is in class $j$ for $i \in [N]$ and $j \in [K]$. If a node $v_i \in \mathcal{V_L}$, then $\mathbf{Y}_{ij}=1$ if the ground truth label of $v_i$ is $j$. If $v_i \notin \mathcal {V_L}$, we initialize $\mathbf{Y}_{ij}=0$ for all $j \in [K]$. Then the soft labels of all the nodes are generated by graph label propagation. %Following the idea of Personalized PageRank (PPR) \cite{klicpera2018predict},
Similar to (\ref{eq:lp}), after $T$ aggregation steps of label propagation, we have $\mathbf{Y}^{(t+1)} = (1 - \beta) \Tilde{\mathbf{A}} \mathbf{Y}^{(t)} + \beta\mathbf{Y}^{(0)}, t=1,...,T-1$,
%\begin{equation}
%\label{eq:propagation}
%        \mathbf{Y}^{(t+1)} = (1 - \beta) %\Tilde{\mathbf{A}} \mathbf{Y}^{(t)} + %\beta\mathbf{Y}^{(0)}, \,\,\, t=1,...,T-1,
%\end{equation}
where $\mathbf{Y}^{(0)} = \mathbf{Y}$, $\beta$ is the teleport probability. The soft labels are then obtained by $\Tilde{\mathbf{Y}} = \text{softmax}(\mathbf{Y}^{(T)})$. We denote the $i$-th row of $\Tilde{\mathbf{Y}}$ by $\tilde{\y}_i$, which is the soft label of node $v_i$.
$f(\cdot)$ is a classifier built by a two-layer MLP followed by a softmax function, which is trained by minimizing the standard loss function for classification, $\LL_{\textup{cls}} = \frac{1}{|\mathcal{V_L}|} \sum_{v_i \in \mathcal{V_L}} H (\tilde{\y}_i, f(\mathbf{h}_{i}))$, where $H$ is the cross-entropy function.
%\begin{equation}
%    \LL_{\textup{cls}} = \frac{1}{|\mathcal{V_L}|} \sum_{v_i \in \mathcal{V_L}} H (\tilde{\y}_i, f(\mathbf{h}_{i})),
%\end{equation}

%In addition to the node classification, we also evaluate the node representation obtained by BRGCL for node clustering. After obtaining the node representations $\left\{\mathbf h_i\right\}_{i=1}^N$, we perform K-means clustering on $\left\{\mathbf h_i\right\}_{i=1}^N$ to obtain node clusters. The number of clusters is set to the ground truth number of classes.
\subsection{Tuning Hyper-Parameters by Cross-Validation}
In this section, we show the tuning procedures on the hyper-parameters $\xi$ from Equation~(\ref{eq:label_update}) and $\gamma_0$ from Equation~(\ref{eq:conf_thres}). We perform cross-validations on $20\%$ of training data to decide the value of $\xi$ and $\gamma_0$. The value of $\xi$ is selected from $\{0.1, 0.15, 0.2, 0.25, 0.3, 0.35, 0.4, 0.45, 0,5\}$. The value of $\gamma_0$ is selected from $\{0.1, 0.2, 0.3, 0.4, 0.5, 0.6, 0.7, 0.8, 0.9\}$. The selected values for $\xi$ and $\gamma_0$ on each dataset are shown in Table~\ref{tab:hyper}.
\begin{table}[h]
\small
	\centering
	\caption{Selected hyper-parameters for each dataset.}
	\vspace{-0.2cm}
\begin{tabular}{cccccc}
\hline
Dataset    & Cora & Citeseer & Pubmed & Coauthor CS & ogbn-arxiv \\ \hline
$\xi$      & 0.20 & 0.15     & 0.35   & 0.40        & 0.25       \\
$\gamma_0$ & 0.3  & 0.5      & 0.7    & 0.4         & 0.4        \\ \hline
\end{tabular}
		
	\label{tab:hyper}
\end{table}
\section{Additional Experiment Results}

\subsection{Detailed Experimental Results with Standard Deviation}
We show detailed experimental results of semi-supervised node classification with symmetric label noise in Table~\ref{table:label_noise_sy}. We run all the experiments for 20 times with random initialization and label noise. Both means and standard deviations of the experiments are shown in Table~\ref{table:label_noise_sy}. Furthermore, detailed experimental results for semi-supervised node classification with attribute noise are shown in Table~\ref{table:feature_noise}.
\label{sec:detail_resutls}
\begin{table}[htbp]
\centering
\caption{Performance comparison on node classification with symmetric label noise. The encoders of methods marked with * are trained with label information.}
\label{table:label_noise_sy}
\scalebox{0.8}{
\begin{tabular}{c|c|ccccc}
\hline
\multirow{2}{*}{Dataset} & \multirow{2}{*}{Methods} & \multicolumn{5}{c}{Noise Level}                                                          \\ \cline{3-7}
                         &                          & 0               & 20              & 40              & 60              & 80              \\ \hline
Cora                     & GCN *                    & $0.817\pm0.005$ & $0.696\pm0.012$ & $0.636\pm0.007$ & $0.517\pm0.010$ & $0.354\pm0.014$ \\
                         & S$^2$GC *                & $\textbf{0.831}\pm\textbf{0.002}$ & $0.711\pm0.012$ & $0.664\pm0.007$ & $0.535\pm0.010$ & $0.396\pm0.014$ \\
                         & GCE                      & $0.819\pm0.004$ & $0.714\pm0.010$ & $0.652\pm0.008$ & $0.509\pm0.011$ & $0.367\pm0.013$ \\
                         & UnionNET                 & $0.820\pm0.006$ & $0.785\pm0.011$ & $0.716\pm0.007$ & $0.594\pm0.010$ & $0.425\pm0.014$ \\
                         & NRGNN                    & $0.822\pm0.006$ & $0.783\pm0.011$ & $0.726\pm0.009$ & $0.590\pm0.014$ & $0.423\pm0.014$ \\
                         & BRGCL                    & $0.822\pm0.006$ & $\textbf{0.793}\pm\textbf{0.009}$ & $\textbf{0.735}\pm\textbf{0.009}$ & $\textbf{0.601}\pm\textbf{0.013}$ & $\textbf{0.446}\pm\textbf{0.012}$ \\ \hline

Citeseer                 & GCN *                    & $0.703\pm0.005$ & $0.624\pm0.008$ & $0.501\pm0.013$ & $0.341\pm0.014$ & $0.301\pm0.019$ \\
                         & S$^2$GC *                & $\textbf{0.727}\pm\textbf{0.005}$ & $0.633\pm0.008$ & $0.528\pm0.013$ & $0.367\pm0.014$ & $0.314\pm0.019$ \\
                         & GCE                      & $0.705\pm0.004$ & $0.629\pm0.008$ & $0.512\pm0.014$ & $0.352\pm0.010$ & $0.305\pm0.014$ \\
                         & UnionNET                 & $0.706\pm0.006$ & $0.643\pm0.012$ & $0.547\pm0.014$ & $0.399\pm0.013$ & $0.332\pm0.013$ \\
                         & NRGNN                    & $0.710\pm0.006$ & $0.659\pm0.008$ & $0.546\pm0.015$ & $0.412\pm0.016$ & $0.369\pm0.018$ \\
                         & BRGCL                    & $0.707\pm0.005$ & $\textbf{0.668}\pm\textbf{0.007}$ & $\textbf{0.569}\pm\textbf{0.013}$ & $\textbf{0.433}\pm\textbf{0.014}$ & $\textbf{0.395}\pm\textbf{0.014}$ \\ \hline

Pubmed                   & GCN *                    & $0.790\pm0.007$ & $0.690\pm0.010$ & $0.574\pm0.012$ & $0.386\pm0.011$ & $0.325\pm0.013$ \\
                         & S$^2$GC *                & $\textbf{0.799}\pm\textbf{0.005}$ & $0.706\pm0.008$ & $0.589\pm0.013$ & $0.401\pm0.014$ & $0.330\pm0.019$ \\
                         & GCE                      & $0.792\pm0.009$ & $0.702\pm0.009$ & $0.581\pm0.011$ & $0.399\pm0.012$ & $0.331\pm0.011$ \\
                         & UnionNET                 & $0.793\pm0.008$ & $0.725\pm0.009$ & $0.623\pm0.012$ & $0.424\pm0.013$ & $0.377\pm0.015$ \\
                         & NRGNN                    & $0.797\pm0.008$ & $0.728\pm0.009$ & $0.618\pm0.013$ & $0.434\pm0.012$ & $0.375\pm0.013$ \\
                         & BRGCL                    & $0.793\pm0.007$ & $\textbf{0.734}\pm\textbf{0.007}$ & $\textbf{0.628}\pm\textbf{0.010}$ & $\textbf{0.458}\pm\textbf{0.010}$ & $\textbf{0.401}\pm\textbf{0.015}$ \\ \hline

Coauthor CS              & GCN *                    & $0.918\pm0.001$ & $0.773\pm0.009$ & $0.656\pm0.006$ & $0.501\pm0.009$ & $0.469\pm0.010$ \\
                         & S$^2$GC *                & $\textbf{0.928}\pm\textbf{0.001}$ & $0.779\pm0.012$ & $0.663\pm0.006$ & $0.514\pm0.009$ & $0.486\pm0.010$ \\
                         & GCE                      & $0.922\pm0.003$ & $0.780\pm0.007$ & $0.659\pm0.007$ & $0.502\pm0.007$ & $0.473\pm0.007$ \\
                         & UnionNET                 & $0.918\pm0.002$ & $0.790\pm0.012$ & $0.671\pm0.013$ & $0.529\pm0.011$ & $0.511\pm0.015$ \\
                         & NRGNN                    & $0.919\pm0.002$ & $0.799\pm0.012$ & $0.689\pm0.009$ & $0.556\pm0.011$ & $0.516\pm0.012$ \\
                         & BRGCL                    & $0.920\pm0.003$ & $\textbf{0.810}\pm\textbf{0.009}$ & $\textbf{0.710}\pm\textbf{0.008}$ & $\textbf{0.572}\pm\textbf{0.011}$ & $\textbf{0.538}\pm\textbf{0.009}$ \\ \hline

ogbn-arxiv               & GCN *                    & $0.717\pm0.003$ & $0.542\pm0.012$ & $0.421\pm0.014$ & $0.336\pm0.011$ & $0.296\pm0.010$ \\
                         & S$^2$GC *                & $\textbf{0.732}\pm\textbf{0.003}$ & $0.553\pm0.011$ & $0.429\pm0.014$ & $0.344\pm0.016$ & $0.306\pm0.013$ \\
                         & GCE                      & $0.720\pm0.004$ & $0.544\pm0.008$ & $0.428\pm0.008$ & $0.344\pm0.019$ & $0.310\pm0.011$ \\
                         & UnionNET                 & $0.724\pm0.006$ & $0.559\pm0.009$ & $0.449\pm0.007$ & $0.367\pm0.008$ & $0.349\pm0.013$ \\
                         & NRGNN                    & $0.721\pm0.006$ & $0.569\pm0.007$ & $0.473\pm0.009$ & $0.379\pm0.008$ & $0.355\pm0.018$ \\
                         & BRGCL                    & $0.720\pm0.005$ & $\textbf{0.579}\pm\textbf{0.007}$ & $\textbf{0.482}\pm\textbf{0.006}$ & $\textbf{0.399}\pm\textbf{0.009}$ & $\textbf{0.376}\pm\textbf{0.015}$ \\ \hline
\end{tabular}
}
\end{table}

\begin{table}[htbp]
\centering
\caption{Performance comparison on node classification with attribute noise. The encoders of methods marked with * are trained with label information.}
\label{table:feature_noise}
\scalebox{0.6}{
\begin{tabular}{c|c|cccccccc}
\hline
\multirow{2}{*}{Dataset} & \multirow{2}{*}{Methods} & \multicolumn{8}{c}{Noise Level}                                                                                                               \\ \cline{3-10}
                         &                          & 0               & 10              & 20              & 30              & 40              & 50              & 60              & 70              \\ \hline
Cora                     & GCN *                    & $0.817\pm0.005$ & $0.785\pm0.012$ & $0.749\pm0.007$ & $0.702\pm0.010$ & $0.639\pm0.014$ & $0.510\pm0.014$ & $0.439\pm0.012$ & $0.378\pm0.014$ \\
                         & S$^2$GC *                & $\textbf{0.831}\pm\textbf{0.002}$ & $0.789\pm0.012$ & $0.760\pm0.007$ & $0.723\pm0.010$ & $0.661\pm0.014$ & $0.521\pm0.014$ & $0.454\pm0.014$ & $0.379\pm0.014$ \\
                         & NRGNN                    & $0.822\pm0.006$ & $0.785\pm0.011$ & $0.753\pm0.007$ & $0.725\pm0.010$ & $0.654\pm0.014$ & $0.517\pm0.014$ & $0.449\pm0.014$ & $0.395\pm0.014$ \\
                         & BRGCL                    & $0.822\pm0.006$ & $\textbf{0.801}\pm\textbf{0.009}$ & $\textbf{0.766}\pm\textbf{0.006}$ & $\textbf{0.740}\pm\textbf{0.010}$ & $\textbf{0.690}\pm\textbf{0.012}$ & $\textbf{0.540}\pm\textbf{0.014}$ & $\textbf{0.469}\pm\textbf{0.014}$ & $\textbf{0.399}\pm\textbf{0.014}$ \\ \hline

Citeseer                 & GCN *                    & $0.703\pm0.005$ & $0.677\pm0.008$ & $0.630\pm0.013$ & $0.601\pm0.014$ & $0.529\pm0.019$ & $0.468\pm0.019$ & $0.372\pm0.019$ & $0.313\pm0.019$ \\
                         & S$^2$GC *                & $\textbf{0.727}\pm\textbf{0.005}$ & $0.695\pm0.008$ & $0.651\pm0.013$ & $0.622\pm0.014$ & $0.553\pm0.019$ & $0.491\pm0.019$ & $0.390\pm0.019$ & $0.322\pm0.019$ \\
                         & NRGNN                    & $0.710\pm0.006$ & $0.683\pm0.008$ & $0.640\pm0.012$ & $0.612\pm0.013$ & $0.540\pm0.018$ & $0.501\pm0.019$ & $0.390\pm0.019$ & $0.317\pm0.019$ \\
                         & BRGCL                    & $0.707\pm0.005$ & $\textbf{0.697}\pm\textbf{0.007}$ & $\textbf{0.660}\pm\textbf{0.010}$ & $\textbf{0.635}\pm\textbf{0.010}$ & $\textbf{0.562}\pm\textbf{0.017}$ & $\textbf{0.514}\pm\textbf{0.019}$ & $\textbf{0.399}\pm\textbf{0.019}$ & $\textbf{0.329}\pm\textbf{0.019}$ \\ \hline

Pubmed                   & GCN *                    & $0.790\pm0.007$ & $0.743\pm0.010$ & $0.702\pm0.012$ & $0.659\pm0.011$ & $0.595\pm0.013$ & $0.532\pm0.013$ & $0.488\pm0.013$ & $0.467\pm0.013$ \\
                         & S$^2$GC *                & $\textbf{0.799}\pm\textbf{0.005}$ & $0.762\pm0.008$ & $0.718\pm0.013$ & $0.679\pm0.014$ & $0.610\pm0.019$ & $0.551\pm0.013$ & $0.497\pm0.013$ & $0.485\pm0.013$ \\
                         & NRGNN                    & $0.797\pm0.008$ & $0.755\pm0.009$ & $0.711\pm0.012$ & $0.670\pm0.013$ & $0.603\pm0.013$ & $0.545\pm0.013$ & $0.499\pm0.013$ & $0.483\pm0.013$ \\
                         & BRGCL                    & $0.793\pm0.007$ & $\textbf{0.770}\pm\textbf{0.007}$ & $\textbf{0.725}\pm0.010$ & $\textbf{0.689}\pm0.010$ & $\textbf{0.625}\pm\textbf{0.015}$ & $\textbf{0.563}\pm\textbf{0.013}$ & $\textbf{0.510}\pm\textbf{0.013}$ & $\textbf{0.501}\pm\textbf{0.013}$ \\ \hline

Coauthor CS              & GCN *                    & $0.918\pm0.001$ & $0.879\pm0.009$ & $0.836\pm0.006$ & $0.791\pm0.009$ & $0.702\pm0.010$ & $0.628\pm0.010$ & $0.531\pm0.010$ & $0.455\pm0.010$ \\
                         & S$^2$GC *                & $\textbf{0.928}\pm\textbf{0.001}$ & $0.890\pm0.009$ & $0.844\pm0.006$ & $0.799\pm0.009$ & $0.713\pm0.010$ & $0.638\pm0.010$ & $0.556\pm0.010$ & $0.476\pm0.010$ \\
                         & NRGNN                    & $0.919\pm0.002$ & $0.885\pm0.012$ & $0.843\pm0.009$ & $0.796\pm0.011$ & $0.710\pm0.012$ & $0.632\pm0.010$ & $0.560\pm0.010$ & $0.469\pm0.010$ \\
                         & BRGCL                    & $0.920\pm0.003$ & $\textbf{0.890}\pm\textbf{0.009}$ & $\textbf{0.849}\pm\textbf{0.008}$ & $\textbf{0.806}\pm\textbf{0.011}$ & $\textbf{0.722}\pm\textbf{0.009}$ & $\textbf{0.653}\pm\textbf{0.010}$ & $\textbf{0.572}\pm\textbf{0.010}$ & $0.488\pm\textbf{0.010}$ \\ \hline

ogbn-arxiv               & GCN *                    & $0.717\pm0.001$ & $0.656\pm0.009$ & $0.609\pm0.006$ & $0.563\pm0.009$ & $0.478\pm0.010$ & $0.402\pm0.010$ & $0.339\pm0.010$ & $0.289\pm0.010$ \\
                         & S$^2$GC *                & $\textbf{0.732}\pm\textbf{0.001}$ & $0.670\pm0.009$ & $0.615\pm0.006$ & $0.570\pm0.009$ & $0.492\pm0.010$ & $0.410\pm0.010$ & $0.344\pm0.010$ & $0.301\pm0.010$ \\
                         & NRGNN                    & $0.721\pm0.002$ & $0.663\pm0.012$ & $0.616\pm0.009$ & $0.577\pm0.011$ & $0.485\pm0.012$ & $0.414\pm0.010$ & $0.342\pm0.010$ & $0.312\pm0.010$ \\
                         & BRGCL                    & $0.720\pm0.003$ & $\textbf{0.682}\pm\textbf{0.009}$ & $\textbf{0.625}\pm\textbf{0.008}$ & $\textbf{0.592}\pm\textbf{0.011}$ & $\textbf{0.495}\pm\textbf{0.009}$ & $\textbf{0.422}\pm\textbf{0.010}$ & $\textbf{0.350}\pm\textbf{0.010}$ & $\textbf{0.330}\pm\textbf{0.010}$ \\ \hline
\end{tabular}
}
\end{table}

\subsection{Asymmetric Label Noise.}
\begin{figure}[tbp]
    \centering
    \subfigure[Cora]{\includegraphics[width=0.3\textwidth]{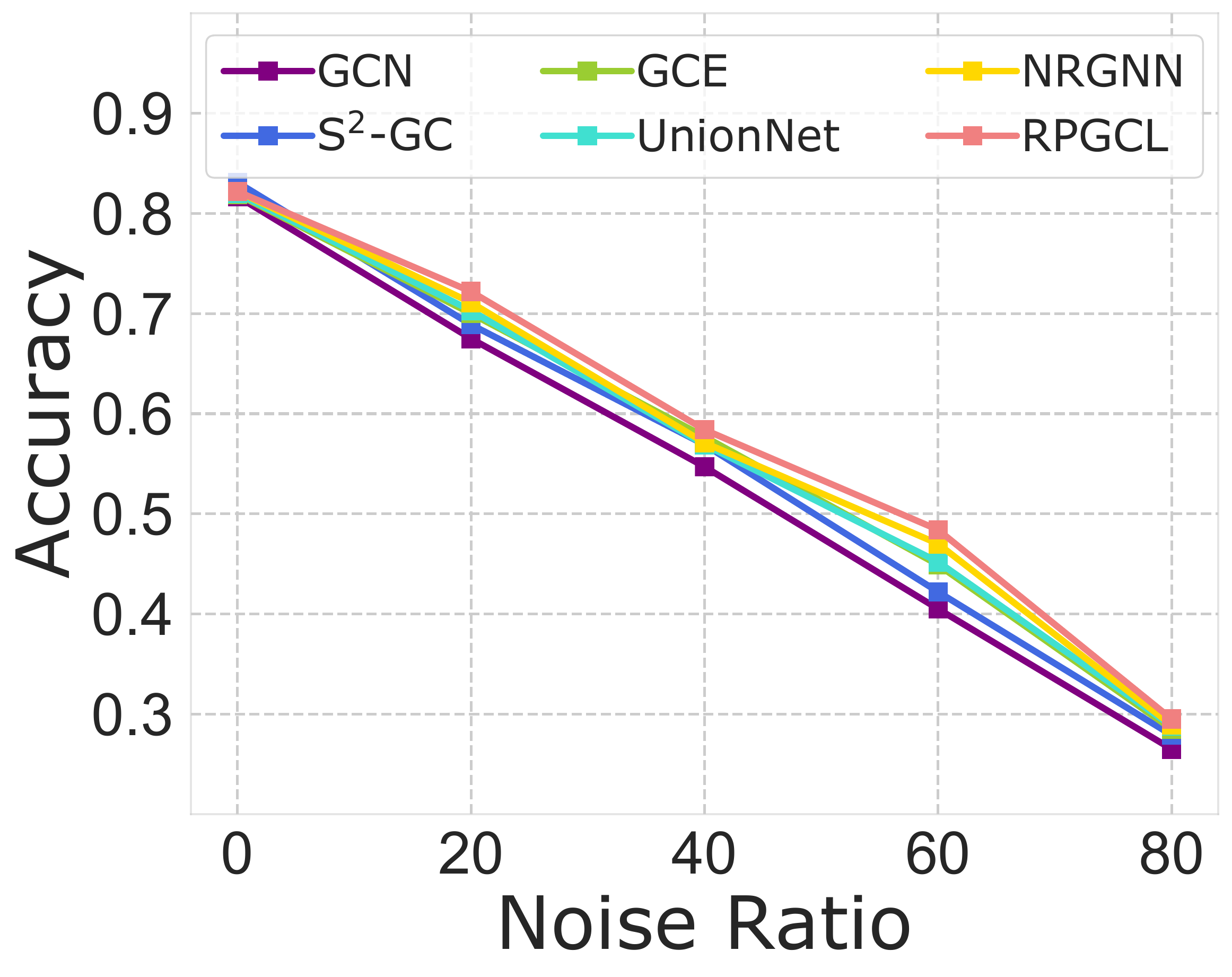}}
    \subfigure[Citeseer]{\includegraphics[width=0.3\textwidth]{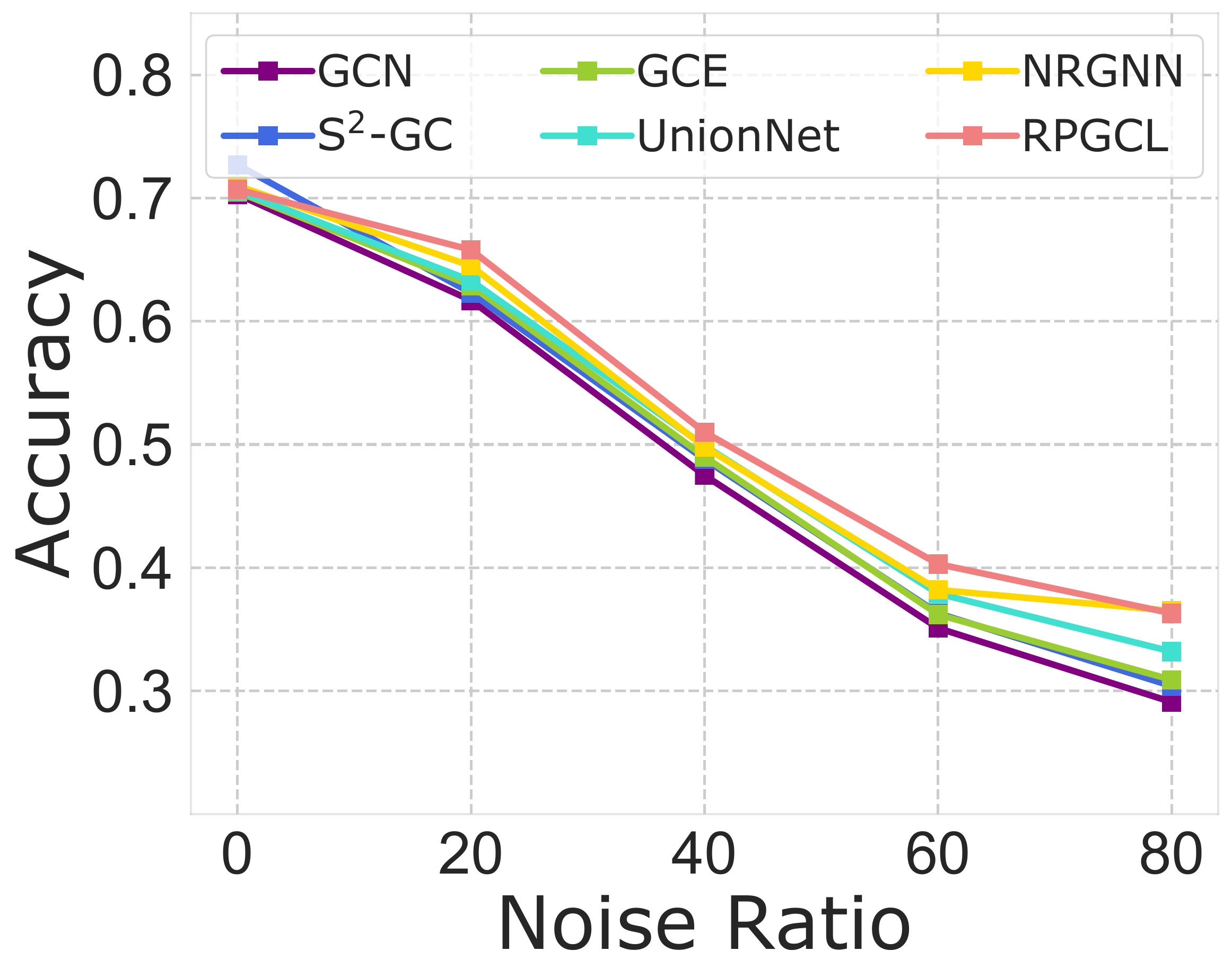}}
    \subfigure[Pubmed]{\includegraphics[width=0.3\textwidth]{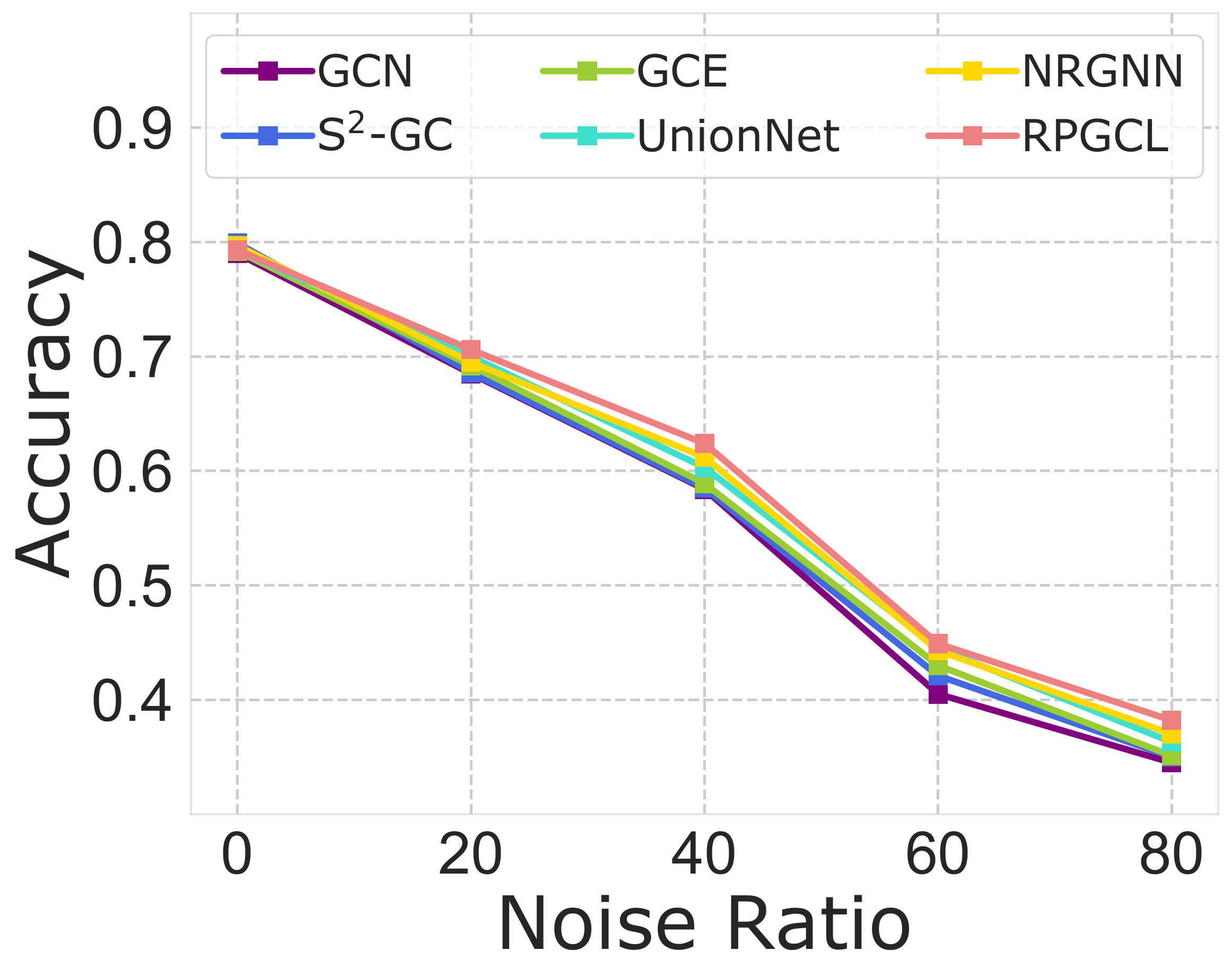}}
    \subfigure[Coauthor CS]{\includegraphics[width=0.3\textwidth]{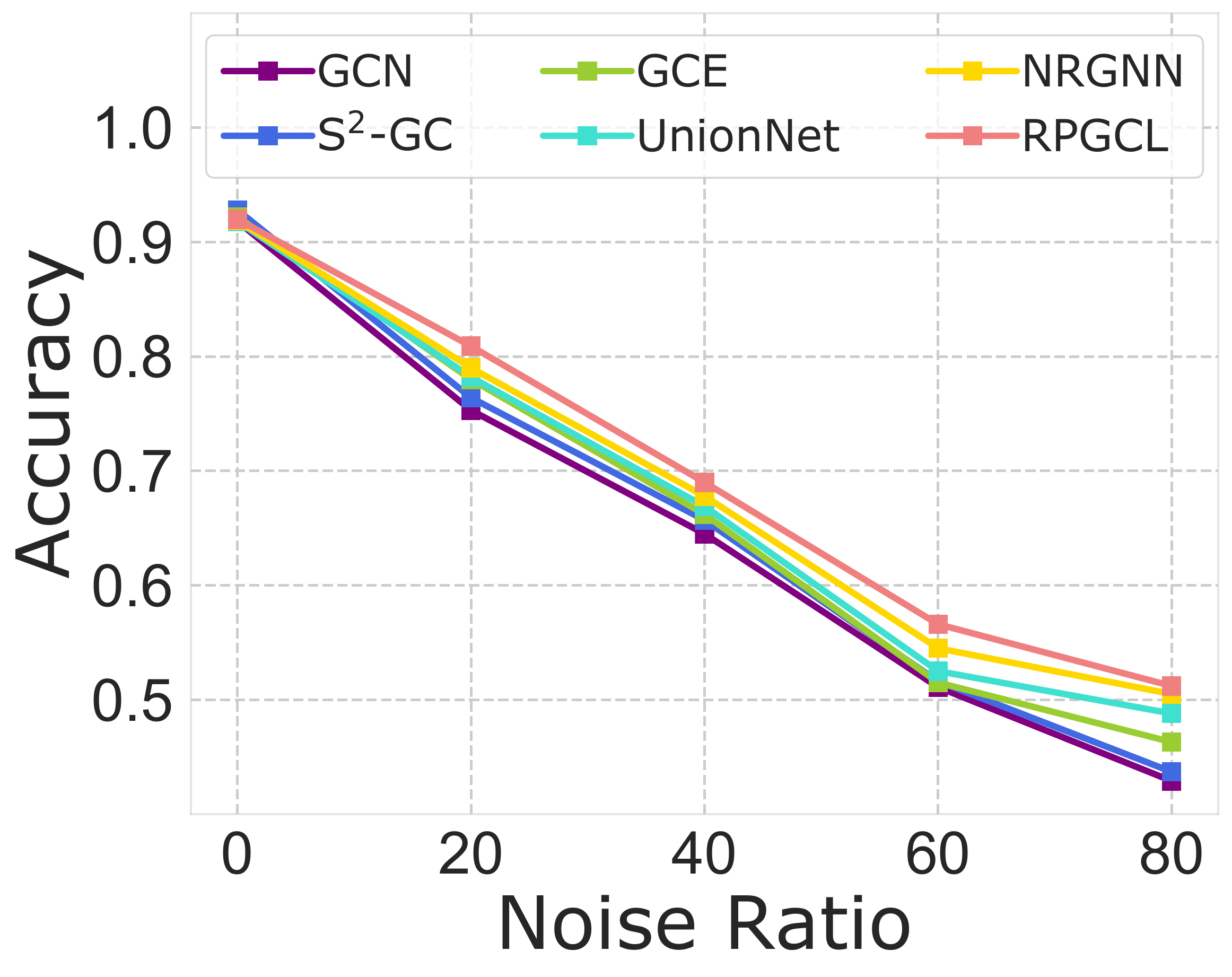}}
    \subfigure[ogbn-arxiv]{\includegraphics[width=0.3\textwidth]{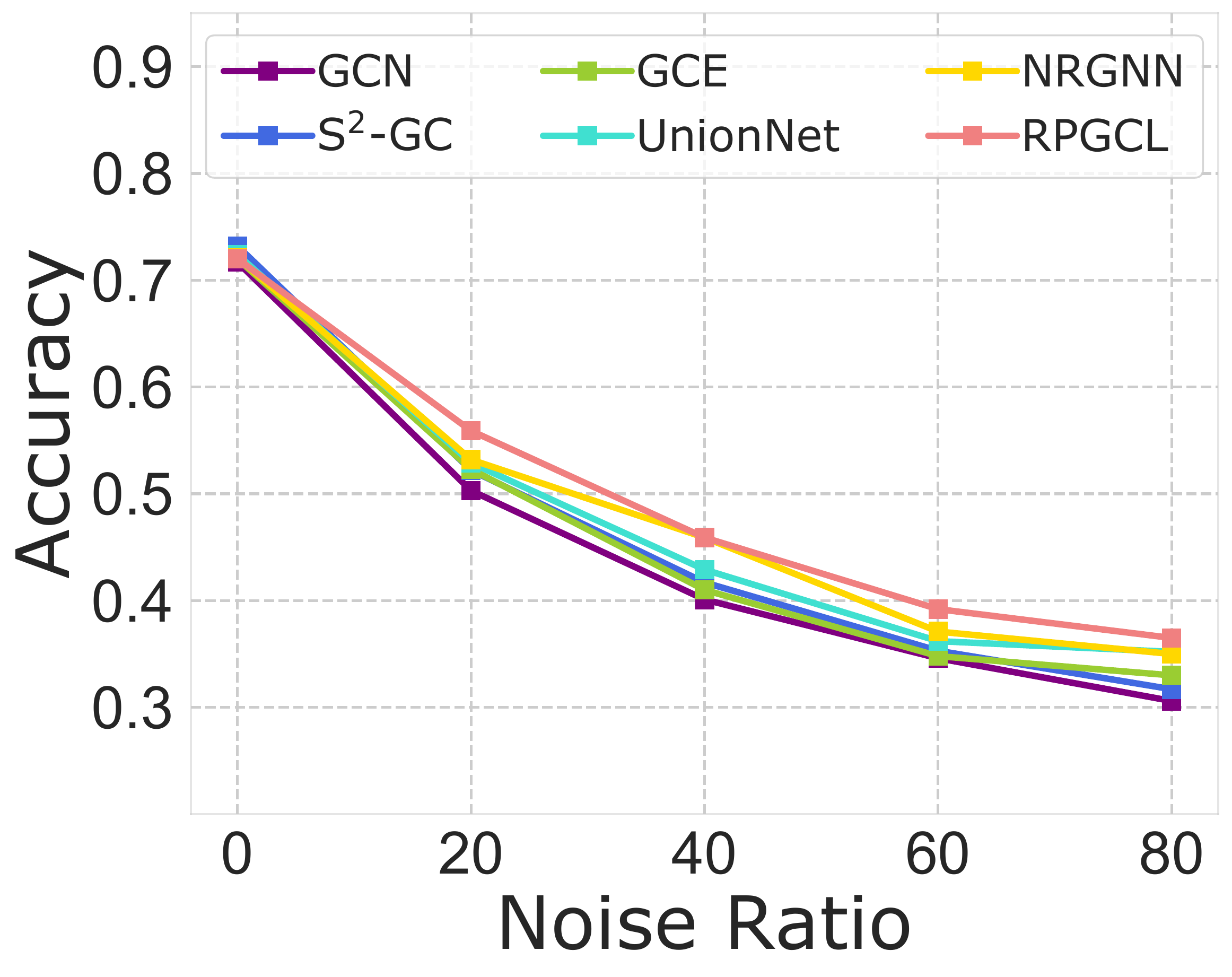}}
    \vspace{-0.1cm}
    \caption{Performance comparisons on semi-supervised node classification with different levels of asymmetric label noise.}
    \label{fig:label_noise_result_asy}
\end{figure}

In addition to the experiments on semi-supervised node classification with symmetric label noise, we also evaluate the performance of BRGCL with different levels of asymmetric label noise. The results are shown in Figure~\ref{fig:label_noise_result_asy}. We run all the experiments for 20 times with random initialization and injected asymmetric label noise. We also report the means and standard deviations of different runs in Table~\ref{table:label_noise_asy}. Similar to the results on semi-supervised node classification with symmetric label noise, BRGCL outperforms competing methods on most levels of asymmetric label noise.
\label{sec:asy_label}
% \subsection{Evaluation Results}

\begin{table}[htbp]
\centering
\caption{Performance comparison on node classification with asymmetric label noise. The encoders of methods marked with * are trained with label information.}
\label{table:label_noise_asy}
\scalebox{0.8}{
\begin{tabular}{c|c|ccccc}
\hline
\multirow{2}{*}{Dataset} & \multirow{2}{*}{Methods} & \multicolumn{5}{c}{Noise Level}                                                                                                                                                   \\ \cline{3-7}
                         &                          & 0                                 & 20                                & 40                                & 60                                & 80                                \\ \hline
Cora                     & GCN *                    & $0.817\pm0.005$                   & $0.675\pm0.012$                   & $0.547\pm0.015$                   & $0.405\pm0.014$                   & $0.265\pm0.012$                   \\
                         & S$^2$GC *                & $\textbf{0.831}\pm\textbf{0.002}$ & $0.689\pm0.012$                   & $0.569\pm0.007$                   & $0.422\pm0.010$                   & $0.279\pm0.014$                   \\
                         & GCE                      & $0.819\pm0.004$                   & $0.700\pm0.010$                   & $0.577\pm0.011$                   & $0.449pm0.011$                    & $0.285\pm0.013$                   \\
                         & UnionNET                 & $0.820\pm0.006$                   & $0.703\pm0.011$                   & $0.569\pm0.014$                   & $0.452\pm0.010$                   & $0.288\pm0.014$                   \\
                         & NRGNN                    & $0.822\pm0.006$                   & $0.711\pm0.011$                   & $0.571\pm0.019$                   & $0.470\pm0.014$                   & $0.289\pm0.022$                   \\
                         & BRGCL                    & $0.822\pm0.006$                   & $\textbf{0.722}\pm\textbf{0.015}$ & $\textbf{0.584}\pm\textbf{0.009}$ & $\textbf{0.484}\pm\textbf{0.013}$ & $\textbf{0.295}\pm\textbf{0.012}$ \\ \hline
Citeseer                 & GCN *                    & $0.703\pm0.005$                   & $0.617\pm0.018$                   & $0.475\pm0.023$                   & $0.351\pm0.014$                   & $0.291\pm0.022$                   \\
                         & S$^2$GC *                & $\textbf{0.727}\pm\textbf{0.005}$ & $0.623\pm0.012$                   & $0.488\pm0.013$                   & $0.363\pm0.012$                   & $0.304\pm0.024$                   \\
                         & GCE                      & $0.705\pm0.004$                   & $0.629\pm0.014$                   & $0.490\pm0.016$                   & $0.362\pm0.015$                   & $0.309\pm0.012$                   \\
                         & UnionNET                 & $0.706\pm0.006$                   & $0.633\pm0.012$                   & $0.499\pm0.015$                   & $0.379\pm0.013$                   & $0.332\pm0.021$                   \\
                         & NRGNN                    & $0.710\pm0.006$                   & $0.645\pm0.012$                   & $0.498\pm0.015$                   & $0.382\pm0.016$                   & $\textbf{0.365}\pm\textbf{0.024}$ \\
                         & BRGCL                    & $0.707\pm0.005$                   & $\textbf{0.658}\pm\textbf{0.017}$ & $\textbf{0.510}\pm\textbf{0.013}$ & $\textbf{0.403}\pm\textbf{0.015}$ & $0.363\pm0.019$                   \\ \hline
Pubmed                   & GCN *                    & $0.790\pm0.007$                   & $0.685\pm0.012$                   & $0.584\pm0.022$                   & $0.405\pm0.025$                   & $0.345\pm0.022$                   \\
                         & S$^2$GC *                & $\textbf{0.799}\pm\textbf{0.005}$ & $0.686\pm0.018$                   & $0.585\pm0.023$                   & $0.421\pm0.030$                   & $0.350\pm0.039$                   \\
                         & GCE                      & $0.792\pm0.009$                   & $0.692\pm0.020$                   & $0.589\pm0.018$                   & $0.430\pm0.012$                   & $0.351\pm0.021$                   \\
                         & UnionNET                 & $0.793\pm0.008$                   & $0.700\pm0.009$                   & $0.603\pm0.020$                   & $0.445\pm0.022$                   & $0.363\pm0.025$                   \\
                         & NRGNN                    & $0.797\pm0.008$                   & $0.695\pm0.009$                   & $0.612\pm0.022$                   & $0.443\pm0.012$                   & $0.370\pm0.023$                   \\
                         & BRGCL                    & $0.793\pm0.007$                   & $\textbf{0.706}\pm\textbf{0.007}$ & $\textbf{0.624}\pm\textbf{0.024}$ & $\textbf{0.449}\pm\textbf{0.020}$ & $\textbf{0.382}\pm\textbf{0.025}$ \\ \hline
Coauthor CS              & GCN *                    & $0.918\pm0.001$                   & $0.753\pm0.012$                   & $0.645\pm0.009$                   & $0.511\pm0.013$                   & $0.429\pm0.032$                   \\
                         & S$^2$GC *                & $\textbf{0.928}\pm\textbf{0.001}$ & $0.764\pm0.022$                   & $0.657\pm0.012$                   & $0.516\pm0.013$                   & $0.437\pm0.020$                   \\
                         & GCE                      & $0.922\pm0.003$                   & $0.780\pm0.015$                   & $0.662\pm0.017$                   & $0.515\pm0.016$                   & $0.463\pm0.017$                   \\
                         & UnionNET                 & $0.918\pm0.002$                   & $0.782\pm0.012$                   & $0.669\pm0.023$                   & $0.525\pm0.011$                   & $0.488\pm0.015$                   \\
                         & NRGNN                    & $0.919\pm0.002$                   & $0.790\pm0.022$                   & $0.678\pm0.014$                   & $0.545\pm0.021$                   & $0.505\pm0.022$                   \\
                         & BRGCL                    & $0.920\pm0.003$                   & $\textbf{0.809}\pm\textbf{0.019}$ & $\textbf{0.690}\pm\textbf{0.012}$ & $\textbf{0.566}\pm\textbf{0.021}$ & $\textbf{0.512}\pm\textbf{0.018}$ \\ \hline
ogbn-arxiv               & GCN *                    & $0.717\pm0.003$                   & $0.503\pm0.022$                   & $0.401\pm0.014$                   & $0.346\pm0.021$                   & $0.306\pm0.022$                   \\
                         & S$^2$GC *                & $\textbf{0.732}\pm\textbf{0.003}$ & $0.522\pm0.021$                   & $0.417\pm0.017$                   & $0.353\pm0.031$                   & $0.317\pm0.023$                   \\
                         & GCE                      & $0.720\pm0.004$                   & $0.523\pm0.019$                   & $0.410pm0.018$                    & $0.348\pm0.019$                   & $0.330\pm0.014$                   \\
                         & UnionNET                 & $0.724\pm0.006$                   & $0.529\pm0.017$                   & $0.429\pm0.021$                   & $0.362\pm0.018$                   & $0.352\pm0.019$                   \\
                         & NRGNN                    & $0.721\pm0.006$                   & $0.532\pm0.018$                   & $\textbf{0.459}\pm\textbf{0.026}$ & $0.371\pm0.020$                   & $0.350\pm0.018$                   \\
                         & BRGCL                    & $0.720\pm0.005$                   & $\textbf{0.559}\pm\textbf{0.020}$ & $\textbf{0.459}\pm\textbf{0.013}$ & $\textbf{0.392}\pm\textbf{0.018}$ & $\textbf{0.365}\pm\textbf{0.015}$ \\ \hline
\end{tabular}
}
\end{table}

\subsection{Node Clustering with Input Attribute Noise}
To further demonstrate the robustness of the node representations learned by BRGCL, we perform node clustering on input with $50\%$ attribute noise. We follow the same evaluation protocol as that in Section~\ref{sec:results}. K-means is then applied to the learned node representations to obtain the clustering results. It can be observed from Table~\ref{tab:clustering_noise} that BRGCL outperforms the competing methods and it is more robust to attribute noise.
\label{sec:clustering_noise}
\begin{table*}[htbp]
\centering
\caption{Node clustering performance comparison on benchmark datasets with $50\%$ input attribute noise.}
\scalebox{0.85}{
\begin{tabular}{cccccccccc}
\hline
\multirow{2}{*}{Methods} & \multicolumn{3}{c}{Cora} & \multicolumn{3}{c}{Citeseer} & \multicolumn{3}{c}{Pubmed} \\ \cline{2-10}
                         & ACC    & NMI    & ARI    & ACC      & NMI     & ARI     & ACC     & NMI     & ARI    \\ \hline
\multicolumn{10}{c}{Supervised}                                                                                 \\ \hline
GCN                      & 64.3   & 48.3   & 50.9   & 64.8     & 37.9    & 40.1    & 64.1    & 30.2    & 27.7   \\
S$^2$GC                  & 65.6   & 50.7   & 51.8   & 65.1     & 37.8    & 40.5    & 65.1    & 30.2    & 28.2   \\
NRGNN                    & 68.1   & 50.6   & 52.1   & 65.3     & 38.6    & 40.7    & 64.9    & 31.2    & 27.8   \\ \hline
\multicolumn{10}{c}{Unsupervised}                                                                               \\ \hline
K-means                  & 45.2   & 29.1   & 22.9   & 50.0     & 28.5    & 26.8    & 55.5    & 31.5    & 28.0   \\
GAE                      & 55.6   & 38.9   & 34.7   & 37.8     & 17.6    & 12.4    & 63.2    & 27.7    & 27.9   \\
ARGA                     & 60.0   & 41.9   & 35.2   & 54.3     & 32.0    & 32.1    & 64.8    & 30.5    & 27.5   \\
ARVGA                    & 60.0   & 42.0   & 37.4   & 51.4     & 26.1    & 23.5    & 64.0    & 29.0    & 28.6   \\
GALA                     & 70.5   & 53.6   & 53.1   & 66.3     & 38.1    & 40.6    & 64.3    & 31.7    & 28.1   \\
MVGRL                    & 70.8   & 53.8   & 53.0   & 66.6     & \textbf{38.7}   & 40.2    & 65.6    & 31.9    & 28.5   \\
BRGCL                     & \textbf{71.2}   & \textbf{54.3}   & \textbf{51.4}   & \textbf{67.1}     & 38.4    & \textbf{41.2}    & \textbf{66.1}    & \textbf{32.1}    & \textbf{29.4}   \\ \hline
\end{tabular}
}
\vspace{-0.15cm}

\label{tab:clustering_noise}
\end{table*}

\subsection{Comparisons to Existing Sample Selection Methods}
\label{sec:sample_selection}
In this subsection, we compare BRGCL against previous sample selection methods including Co-teaching~\cite{Han2018NIPS} and Self-Training~\cite{li2018deeper} for node classification with symmetric label noise. Co-teaching maintains two networks to select clean samples for each other. Self-Training finds nodes with the most confident pseudo labels, and it augmented the labeled training data by incorporating confident nodes with their pseudo labels into the existing training data. The results are shown in Table~\ref{table:label_noise_sample}. We can clearly see that BRGCL greatly outperforms competing sample selection methods.
\begin{table}[htbp]
\centering
\caption{Performance comparison againist previous sample selection methods on node classification with different levels of symmetric label noise.}
\label{table:label_noise_sample}
\scalebox{0.8}{
\begin{tabular}{c|c|ccccc}
\hline
\multirow{2}{*}{Dataset} & \multirow{2}{*}{Methods} & \multicolumn{5}{c}{Noise Level}                                                                                                                                                   \\ \cline{3-7}
                         &                          & 0                                 & 20                                & 40                                & 60                                & 80                                \\ \hline
Cora                     & Self-training            & $0.820\pm0.005$                   & $0.743\pm0.014$                   & $0.684\pm0.007$                   & $0.562\pm0.013$                   & $0.414\pm0.012$                    \\
                         & Co-teaching              & $\textbf{0.822}\pm\textbf{0.005}$ & $0.722\pm0.010$                   & $0.695\pm0.009$                   & $0.557\pm0.010$                   & $0.402\pm0.017$                   \\
                         & BRGCL                    & $\textbf{0.822}\pm\textbf{0.006}$ & $\textbf{0.793}\pm\textbf{0.009}$ & $\textbf{0.735}\pm\textbf{0.006}$ & $\textbf{0.601}\pm\textbf{0.010}$ & $\textbf{0.446}\pm\textbf{0.012}$ \\ \hline
Citeseer                 & Self-training            & $0.706\pm0.005$                   & $0.654\pm0.018$                   & $0.541\pm0.014$                   & $0.397\pm0.013$                   & $0.341\pm0.022$                   \\
                         & Co-teaching              & $0.705\pm0.005$                   & $0.648\pm0.014$                   & $0.518\pm0.017$                   & $0.373\pm0.014$                   & $0.319\pm0.020$                   \\
                         & BRGCL                    & $\textbf{0.707}\pm\textbf{0.005}$ & $\textbf{0.668}\pm\textbf{0.007}$ & $\textbf{0.569}\pm\textbf{0.010}$ & $\textbf{0.433}\pm\textbf{0.010}$ & $\textbf{0.395}\pm\textbf{0.017}$ \\ \hline
Pubmed                   & Self-training            & $0.792\pm0.007$                   & $0.722\pm0.015$                   & $0.597\pm0.019$                   & $0.419\pm0.021$                   & $0.375\pm0.023$                   \\
                         & Co-teaching              & $0.790\pm0.005$                   & $0.720\pm0.012$                   & $0.584\pm0.013$                   & $0.403\pm0.014$                   & $0.342\pm0.022$                   \\
                         & BRGCL                    & $\textbf{0.793}\pm\textbf{0.005}$ & $\textbf{0.734}\pm\textbf{0.007}$ & $\textbf{0.628}\pm\textbf{0.010}$ & $\textbf{0.458}\pm\textbf{0.010}$ & $\textbf{0.401}\pm\textbf{0.015}$ \\ \hline
Coauthor CS              & Self-training            & $0.918\pm0.004$                   & $0.784\pm0.006$                   & $0.672\pm0.010$                   & $0.542\pm0.013$                   & $0.497\pm0.015$                   \\
                         & Co-teaching              & $\textbf{0.921}\pm\textbf{0.003}$ & $0.780\pm0.009$                   & $0.666\pm0.012$                   & $0.529\pm0.015$                   & $0.488\pm0.019$                   \\
                         & BRGCL                    & $0.920\pm0.003$                   & $\textbf{0.810}\pm\textbf{0.009}$ & $\textbf{0.710}\pm\textbf{0.008}$ & $\textbf{0.572}\pm\textbf{0.011}$ & $\textbf{0.538}\pm\textbf{0.009}$ \\ \hline
ogbn-arxiv               & Self-training            & $\textbf{0.722}\pm\textbf{0.003}$ & $0.572\pm0.011$                   & $0.462\pm0.012$                   & $0.368\pm0.018$                   & $0.336\pm0.020$                   \\
                         & Co-teaching              & $0.718\pm0.006$                   & $0.559pm0.017$                    & $0.437\pm0.024$                   & $0.359\pm0.016$                   & $0.322\pm0.025$                   \\
                         & BRGCL                    & $0.720\pm0.005$                   & $\textbf{0.579}\pm\textbf{0.007}$ & $\textbf{0.482}\pm\textbf{0.006}$ & $\textbf{0.399}\pm\textbf{0.009}$ & $\textbf{0.376}\pm\textbf{0.015}$ \\ \hline
\end{tabular}
}
\end{table}

\begin{table}[htbp]
\caption{Ablation study on contrastive components for node classification with label noise.}
\centering
\scalebox{0.9}{
\begin{tabular}{ccccccc}
\hline
\multirow{2}{*}{Method} & \multicolumn{2}{c}{Cora}          & \multicolumn{2}{c}{Citeseer}      & \multicolumn{2}{c}{Pubmed}        \\ \cline{2-7}
                        & Confident       & K-means            & Confident       & K-means            & Confident       & K-means            \\ \hline
Joint                   & 77.7$\pm{0.08}$ & 78.2$\pm{0.11}$ & 65.7$\pm{0.09}$ & 65.3$\pm{0.10}$ & 72.5$\pm{0.10}$ & 72.5$\pm{0.12}$ \\
Decoupled               & 79.3$\pm{0.09}$ & 78.5$\pm{0.09}$ & 66.8$\pm{0.07}$ & 66.3$\pm{0.08}$ & 73.4$\pm{0.07}$ & 73.0$\pm{0.09}$ \\ \hline
\end{tabular}
	}

\label{tab:ablation}
\end{table}
\subsection{Joint Training vs. Decoupled Training.}
\label{sec:decoupled}
We study the effectiveness of our decoupled training framework compared with jointly training the encoder and the classifier. We compare the performance on node classification with $20\%$ label noise level. The results are shown in Table~\ref{tab:ablation}. It can be observed that decoupling the training of classifier and encoder can mitigate the effects of label noise. %In addition, we also study the strength of the confident prototype compared with prototypes estimated by K-means following \cite{PCL}. The results clearly show that confident prototypes can boost the performance of down stream tasks.

\subsection{Number of Confident Prototypes.}
\label{sec:number_proto}

To further study the behavior of BRGCL, we show the number of robust prototypes estimated by BPL in Table~\ref{tab:proto}. It can be observed from the results that the estimated number of robust prototypes is usually very close to the ground truth number of classes for different datasets, justifying the effectiveness of BPL. Because BEC is based on the pseudo labels estimated by BPL, the success of BPL leads to trustworthy estimation of confident nodes and robust prototypes by BEC.

\begin{table}[H]
	\small
	\centering
		\caption{Number of robust prototypes inferred by BPL compared with the ground truth number of classes on different datasets }
	\scalebox{1.0}{

\begin{tabular}{cccccc}
\hline
Datasets & Citeseer & Cora & Pubmed & Coauthor CS & ogbn-arxiv \\
\hline
$\xi$ in eq. (\ref{eq:label_update})         & 0.15       & 0.20     & 0.35  & 0.30     & 0.4\\
Estimated $K$   & 6          & 8        & 3     & 19& 48\\
Classes         & 6          & 7        & 3     &15 & 40\\
\hline
\end{tabular}

}

	\label{tab:proto}
\end{table}

\end{document}